\documentclass[journal]{IEEEtran}
\makeatletter
\let\NAT@parse\undefined
\makeatother
\usepackage[colorlinks]{hyperref}
\hypersetup{colorlinks=true,linkcolor=red,citecolor=blue,urlcolor=magenta}

\usepackage[utf8]{inputenc}
\usepackage[T1]{fontenc}
\usepackage{graphicx}
\usepackage{multirow}
\usepackage{amsmath}
\usepackage{amssymb} 
\usepackage{verbatim}
\usepackage{makecell}
\usepackage{caption}
\usepackage{pifont}
\usepackage{booktabs}
\usepackage[table,dvipsnames]{xcolor} % for rowcolor

\usepackage{amsmath,amssymb}
\usepackage[ruled,linesnumbered]{algorithm2e}

\begin{document}

\title{\LARGE\bf TRACER: Texture-Robust Affordance Chain-of-Thought for Deformable-Object Region Grounding
}

\author{Anonymous Authors}
\author{Wanjun Jia$^{1}$, Kang Li$^{1}$, Fan Yang$^{1}$, Mengfei Duan$^{1}$, Wenrui Chen$^{1,2}$,\\Yiming Jiang$^{1,2}$, Hui Zhang$^{1,2}$, Kailun Yang$^{1,2,*}$, Zhiyong Li$^{1,2,*}$, and Yaonan Wang$^{1,2}$%
\thanks{This work was partially supported by
the National Natural Science Foundation of China under Grant No. 62273137, No. 62473139, No. U21A20518, No. U23A20341, and No. 62388101, the Hunan Provincial Research and Development Project under Grant 2025QK3019, the Hunan Science Fund for Distinguished Young Scholars under Grant 2024JJ2027, and the State Key Laboratory of Autonomous Intelligent Unmanned Systems (the opening project number ZZKF2025-2-10). 
}
\thanks{$^{1}$The authors are with the School of Artificial Intelligence and Robotics, Hunan University, Changsha 410012, China. (E-mail: kailun.yang@hnu.edu.cn, zhiyong.li@hnu.edu.cn.)}%
\thanks{$^{2}$The authors are also with the National Engineering Research Center of Robot Visual Perception and Control Technology, Hunan University, Changsha 410082, China.}%
\thanks{$^{*}$Corresponding authors: Kailun Yang and Zhiyong Li.}
}

\maketitle

\begin{abstract}
The central challenge in robotic manipulation of deformable objects lies in aligning high-level semantic instructions with physical interaction points under complex appearance and texture variations. Existing vision-based affordance prediction methods often suffer from boundary overflow and fragmented functional regions. To address these issues, we propose TRACER, a Texture-Robust Affordance Chain-of-Thought for Deformable-Object Region Grounding framework that maps hierarchical semantic reasoning to appearance-robust and physically consistent interaction regions. Specifically, a Tree-structured Affordance Chain-of-Thought (TA-CoT) decomposes high-level task intentions into hierarchical affordance-semantic instructions. A Spatially-Constrained Boundary Refinement (SCBR) mechanism suppresses prediction spillover and guides responses toward valid object regions. Furthermore, an Interactive Convergent Refinement Flow (ICRF) refines dispersed affordance responses into coherent regions, improving spatial continuity and physical plausibility. Experiments on the Fine-AGDDO15 dataset and a real-world robotic platform demonstrate that TRACER improves affordance grounding precision across diverse textures and patterns. It also enhances long-horizon manipulation success, bridging high-level semantic reasoning and low-level physical execution. The source code and dataset will be made publicly available at \url{https://github.com/Dikay1/TRACER}.
\end{abstract}

\renewcommand{\abstractname}{Note to Practitioners}
\begin{abstract}
This work addresses reliable affordance region grounding for deformable-object manipulation under complex textures, large deformations, and ambiguous visual cues. Existing vision-based methods often struggle to identify stable interaction regions when object appearance changes significantly or when texture interference causes fragmented and spatially inconsistent responses. Our framework combines hierarchical semantic reasoning with spatial constraints and flow-based refinement to generate coherent functional regions from limited training data. These grounded regions are then converted into manipulation points to support downstream robotic tasks such as folding, hanging, and object extraction. Experiments on a dual-arm robotic platform demonstrate improved grounding robustness and higher manipulation success across different object appearances and task settings. A current limitation is that the framework mainly relies on visual perception and lacks explicit feedback about contact states during manipulation. Future work will incorporate tactile sensing and adaptive control to improve robustness under uncertain contact, severe deformation, and more challenging real-world conditions.
\end{abstract}

\begin{IEEEkeywords}
Deformable Object Manipulation, Affordance Grounding, Tree-structured Chain-of-Thought.
\end{IEEEkeywords}

\section{Introduction}

\IEEEPARstart{W}{ith} the continuous evolution of embodied artificial intelligence, the application of service robots in long-horizon tasks, such as household organization, has garnered significant attention~\cite{lu2024garmentlab}.
Unlike rigid objects, deformable objects in household scenarios (\textit{e.g.}, clothes, hats, and towels) exhibit near-infinite degrees of freedom, complex physical dynamics, and intricate textural distributions~\cite{borras2020grasping,wu2025garmentpile,zhou2025learning,10966003}. 
These attributes pose dual challenges~\cite{wu2024unigarmentmanip, wu2023learning}: generating semantic guidance that respects both topological constraints and task sequence, and achieving precise pixel-level affordance grounding amidst intense textural interference.

\begin{figure}[t!]
\centerline{\includegraphics[width=0.48\textwidth]{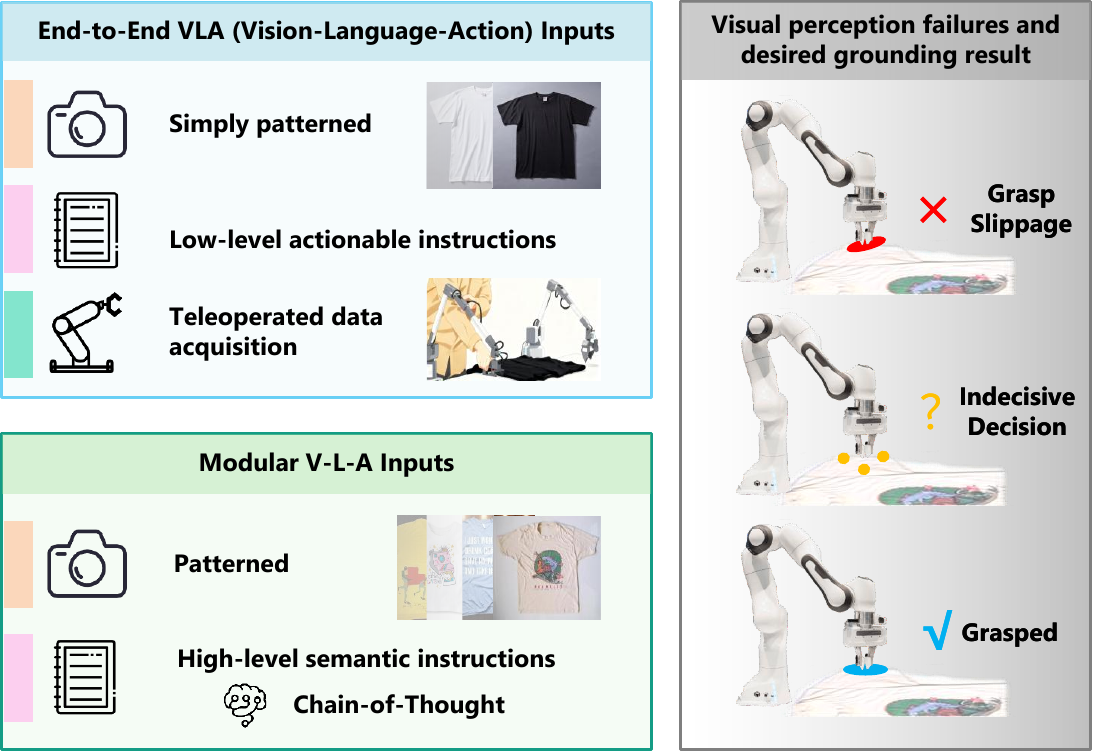}}
\captionsetup{font=small}
\caption{\small Comparison of manipulation paradigms and failure analysis for deformable objects. Blue shows end-to-end VLA inputs; green shows modular V-L-A inputs with CoT. Gray highlights common perception failures versus desired physically consistent grounding.
}
\label{intro}
\vskip-2.2ex
\end{figure}

As illustrated in Fig.~\ref{intro}, current research paradigms for deformable object manipulation can be broadly categorized into two streams. 
The first paradigm is represented by end-to-end Vision-Language-Action (VLA) models (blue region in Fig.~\ref{intro}), which jointly model language, vision, and action through large-scale action data collection. 
Although such models achieve promising performance in partially controlled settings, their prohibitive data acquisition costs and substantial computational requirements severely limit their practicality in real-world deployment. 
More critically, most existing benchmarks are constructed using simplified deformable objects with uniform colors and simple geometries (\textit{e.g.}, ropes~\cite{zhang2024adaptigraph, caporali2025gnn} or uniform-color cloths~\cite{zhuang2025flat, canberk2022cloth}). 
When transferred to real-world household environments containing deformable objects with complex patterns (\textit{e.g.}, plaid shirts or printed towels), the underlying perceptual networks frequently suffer from feature confusion, leading to catastrophic prediction failures.

To enhance data efficiency and the transparency of task reasoning, another stream of research has shifted toward modular V-L-A frameworks~\cite{yang2024task,yang2025learning,jia2025one},
as illustrated in the green region of Fig.~\ref{intro}. 
These approaches increasingly leverage the intrinsic Chain-of-Thought (CoT) capabilities of Vision-Language Models (VLMs) for high-level task reasoning. This enables robots to recursively decompose abstract, 
long-horizon instructions (\textit{e.g.}, ``fold T-shirt'') into a sequence of semantically explicit sub-tasks (\textit{e.g.}, ``grasp sleeve''), significantly enhancing reasoning transparency at the semantic level. However, a significant perceptual grounding gap persists between high-level semantic logic and low-level physical execution~\cite{park2025trace, song2025robospatial}.
More severely, existing research on deformable object manipulation predominantly concentrates on decision and execution methodologies, including control strategies~\cite{clegg2020learning, sunil2025reactive}, dynamics and shape modeling~\cite{tian2025diffusion,du2025physics}, and reinforcement learning~\cite{kerr2025eye}, often operating under the idealized assumption that the visual system provides stable and precise interaction priors.
However, in real-world household scenarios, this perceptual premise rarely holds, leading to a noticeable misalignment between high-level semantic reasoning and complex visual inputs.
This grounding difficulty~\cite{pan2025omnimanip, zhao2025cot} manifests in actual manipulation as two key defects, as illustrated in the gray region of Fig.~\ref{intro}:
(1) spatial prediction overflow, where interaction points drift beyond object boundaries into the background, inducing grasp slippage; and (2) functional region fragmentation, where a single functional zone is split into discrete responses, hindering reliable manipulation target selection during physical execution.

To address these challenges, we propose a one-shot long-horizon perception framework termed \textbf{TRACER}, which stands for \textbf{T}exture-Robust \textbf{A}ffordance \textbf{C}hain-of-thought for d\textbf{E}formable-object \textbf{R}egion Grounding. TRACER establishes a cross-hierarchical mapping pathway from hierarchical semantic reasoning to appearance-robust and physically consistent functional region grounding, thereby constructing a perceptual closed-loop that facilitates the transition from high-level task reasoning to low-level physical execution. The framework consists of three synergistic components:

First, to resolve logical reasoning and state dependency issues in long-horizon tasks, we formulate the Tree-structured Affordance Chain-of-Thought, termed TA-CoT. It formalizes previously unstructured complex tasks into a hierarchical decision tree characterized by temporal logic and, through a visual state verification mechanism, ensures that every reasoning step is grounded in the object's actual topological state (\textit{e.g.}, whether the sleeve is in position), thereby achieving physical plausibility in semantic reasoning.
Second, to mitigate grasp slippage induced by spatial prediction overflow, we design the Spatially-Constrained Boundary Refinement (SCBR) loss. By encouraging the model to emphasize global structural coherence rather than local textural variations, the affordance responses are guided to naturally converge within physically valid object boundaries, substantially improving operational robustness and safety.
Finally, to resolve functional region fragmentation and selection uncertainty, we establish the Interactive Convergent Refinement Flow (ICRF). By simulating a physically consistent dynamical convergence process, this module dynamically aggregates loose, multi-modal initial predictions into topologically connected and semantically explicit interaction zones. This achieves pixel-level aggregation from coarse heatmaps to precise execution points, ensuring a stable manipulation target.
Additionally, we employ an affordance region supervision strategy grounded in realistic interaction logic, which constrains model predictions to satisfy physical manipulation requirements in both spatial consistency and functional semantics.

To comprehensively evaluate the proposed approach in real-world scenarios, we evaluate our approach on Fine-AGDDO15, a dataset covering $15$ object categories and $15$ affordance classes, and a physical bimanual robot. TRACER significantly outperforms the OS-AGDO baseline, improving KLD, SIM, and NSS by $4.8\%$, $7.5\%$, and $4.3\%$, respectively. 
In four real-world tasks, TRACER consistently outperforms OS-AGDO, achieving up to $85\%$ success on ``Hang hat'' and $80\%$ on ``Pull out tissue''.
These results substantiate TRACER’s superiority in precise affordance grounding and long-horizon stability against varied visual appearances.

Our contributions can be summarized as follows:

\begin{itemize}
    \item We propose TRACER, the first one-shot long-horizon affordance grounding framework for complex-textured deformable objects, bridging the gap between high-level semantic reasoning and low-level physical execution without requiring expensive action data.
    \item To mitigate prediction overflow and functional region fragmentation in deformable object perception, we introduce two key modules: SCBR and ICRF. By leveraging spatial constraints and dynamic convergence, these modules refine coarse, discrete semantic responses into continuous, physically consistent interaction manifolds.
    \item We conduct multi-dimensional evaluations on the Fine-AGDDO15 dataset and a physical robotic platform, which quantitatively and qualitatively substantiate the robustness of TRACER against diverse textural interferences and its efficacy in significantly enhancing affordance grounding precision and execution stability.
\end{itemize}

\section{Related Work}

\subsection{Robotic Manipulation of Deformable Objects}
Robotic manipulation of deformable objects remains a long-standing challenge. Existing approaches include physics-based modeling methods (\textit{e.g.}, Finite Element Models (FEM), Mass-Spring models) and model-free methods (\textit{e.g.}, Reinforcement Learning, Imitation Learning~\cite{10602544}), which primarily focus on control strategies or dynamics modeling for tasks such as folding~\cite{huang2024rekep, sunil2025reactive}, hanging~\cite{chen2023learning, chen2025graphgarment}, dressing~\cite{zhang2022learning, sun2024force}. 
They often assume that an accurate perception module (\textit{e.g.}, keypoints or object state) already exists, or they rely on simplified deformable objects (\textit{e.g.}, ropes~\cite{zhang2024adaptigraph, caporali2025gnn}, uniform-color towels~\cite{weng2022fabricflownet, jiang2025phystwin}, uniform-color garments~\cite{zhuang2025flat, canberk2022cloth}). 
Recently, end-to-end Vision-Language-Action (VLA) models (\textit{e.g.}, $\pi_0$~\cite{black2410pi0}, $\pi_{0.5}$~\cite{intelligence2504pi05}, and WALL-OSS~\cite{zhai2025igniting}) directly map language instructions to actions via large-scale data-driven learning. However, these methods require costly action data and computation, and typically entangle perception and manipulation in a black-box manner without explicit affordance grounding, which hampers their generalization to real-world tidying scenarios with complex textures and self-occlusion.

To alleviate these challenges, recent work has explored state estimation and representation learning for deformable objects. Self-supervised keypoint detection~\cite{hou2024key} and dense correspondence learning have been used to localize unlabeled semantic points. Simultaneously, generative dynamics models, such as diffusion models and Graph Neural Networks (GNNs), have been introduced for long-horizon prediction and reconstruction under severe self-occlusion, often leveraging simulation environments~\cite{zhou2023clothesnet, wang2025dexgarmentlab} to bridge the gap between simulation and the real world. 
Nevertheless, these efforts remain largely focused on control and dynamics, assuming the perception problem is either solved or simplified. 
In contrast, our work explicitly targets this bottleneck by delivering accurate interaction regions aligned with physical feasibility in real-world tidying scenarios, thus forming a robust connection between explicit affordance grounding and downstream control.

\begin{figure}[t!]
\centerline{\includegraphics[width=0.48\textwidth]{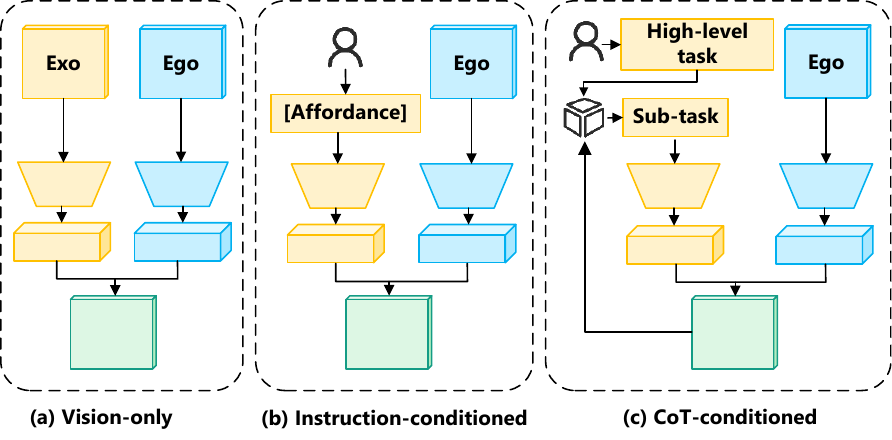}}
\captionsetup{font=small}
\caption{\small Affordance grounding under different semantic conditions. (a) Vision-only: Prediction based only on visual input. (b) Instruction-conditioned: Guided by a single-step language instruction. (c) CoT-conditioned (Ours): Conditioned on hierarchical sub-task semantics from TA-CoT. Ego: first-person view; Exo: third-person view.
}
\label{text_affordance_com}
\vskip-2ex
\end{figure}

\subsection{Semantic-driven Visual Affordance Grounding}
Although Large Language Models (LLMs) have improved robotic high-level reasoning and task planning, a critical grounding gap remains when translating abstract language instructions into concrete physical interaction regions. While Chain-of-Thought (CoT) reasoning enhances the interpretability of task decomposition, its connection to low-level spatial perception is still weak~\cite{pan2025omnimanip, zhao2025manipbench}. As a result, robots often fail to reliably map ambiguous instructions to physically executable regions on deformable objects~\cite{song2025robospatial,ahn2022can}.
To bridge semantic reasoning and spatial execution, recent work has turned to visual affordance learning. Affordance-R1~\cite{wang2025affordancer1} targets generalizable affordance reasoning, while RAGNet~\cite{wu2025ragnet} focuses on reasoning-based affordance segmentation. As shown in Fig.~\ref{text_affordance_com}, existing approaches can be grouped by their degree of semantic conditioning. Vision-only methods (Fig.~\ref{text_affordance_com}(a)) infer action regions from appearance cues or exocentric–egocentric correspondence using heatmap regression or keypoint detection. However, without semantic constraints, they are sensitive to texture and background variations and lack spatial consistency.

To incorporate explicit task intent, instruction-conditioned methods (Fig.~\ref{text_affordance_com}(b)) have been proposed. For example, Huang~\textit{et al.}~\cite{huang2024rekep}, combine multi-modal models like VLMs or DINOv2~\cite{jose2024dinov2meetstextunified}, using semantic features to guide manipulation point prediction and candidate filtering, thereby achieving context-sensitive affordance understanding in semantic space~\cite{sunil2025reactive}. Nevertheless, since language instructions are typically modeled as holistic conditions, these methods struggle to capture structured semantic dependencies within long-horizon tasks or complex manipulation sequences.
Building on this trend, we propose a CoT-conditioned affordance paradigm (Fig.~\ref{text_affordance_com}(c)) that explicitly links hierarchical semantic reasoning with spatially constrained affordance refinement, enabling precise grounding from high-level intentions to concrete interaction regions.

\begin{figure*}[t!]
\centerline{\includegraphics[width=0.9\textwidth]{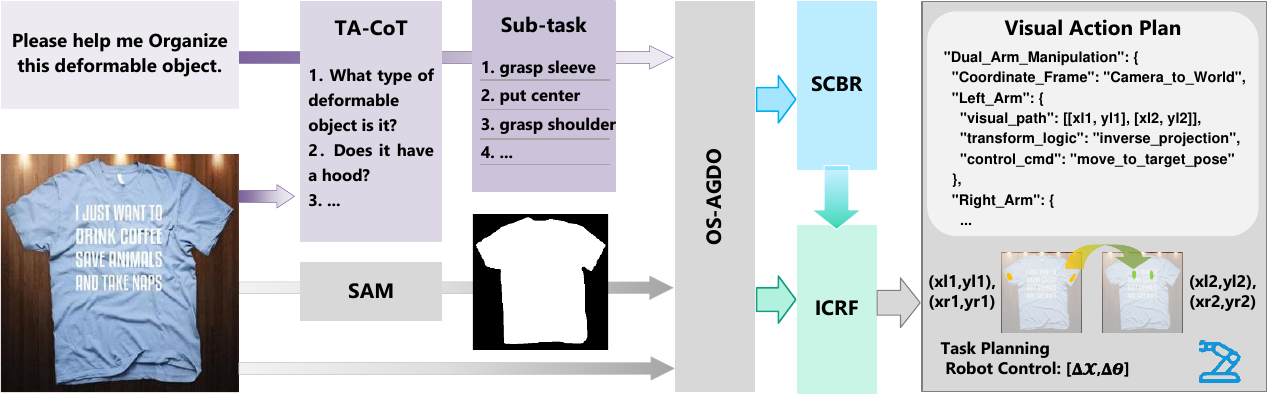}}
\captionsetup{font=small}
\caption{\small Overview of the proposed TRACER framework. TA-CoT provides hierarchical semantic guidance, while SCBR and ICRF progressively refine affordance predictions for closed-loop robotic execution.
}
\label{pipline}
\vskip-2ex
\end{figure*}

\subsection{Generative Modeling and Interaction Field Refinement}
To mitigate the multimodality and instability of semantically-driven affordance distributions, recent studies have introduced dynamic generative modeling paradigms, such as the widely used diffusion models~\cite{tian2025diffusion} and the further developed flow matching methods~\cite{gao2025vita, liu2022flow}.
By modeling distribution transformations in continuous latent spaces, these approaches enable controllable and smooth convergence from initial to target states, and have achieved notable success in dense prediction tasks such as semantic segmentation and depth estimation~\cite{noobject}. However, conventional rectified flow typically relies on a velocity-field-based Mean Squared Error (MSE) objective, which tends to learn averaged motion patterns and struggles to represent the inherent multimodality of affordance distributions. To address this limitation, Hierarchical Rectified Flow (HRF)~\cite{zhang2025towards} introduces multi-level dynamic modeling to more fully characterize latent evolution, leading to more stable and physically consistent convergence.

In the research field of affordance perception, the introduction of generative modeling is still in its nascent stages. AffordDexGrasp~\cite{wei2025afforddexgrasp} adopts a cascaded flow to generate affordance maps and grasp poses, while H2OFlow~\cite{zhang2025h2oflow} points out that most existing methods focus on static predictions and neglect dynamic attributes such as directionality and spatial occupancy. Therefore, this work proposes an interactive convergent refinement flow that explicitly models the pixel-wise evolution of interaction regions, progressively transforming coarse predictions into physically consistent manipulation targets. This enables robust fine-grained affordance grounding for deformable object manipulation.

\section{Methodology}

We propose TRACER, a one-shot long-horizon affordance region grounding framework termed Texture-Robust Affordance Chain-of-thought for dEformable-object Region Grounding, designed for complex-textured deformable objects. The framework establishes a cross-hierarchical mapping from hierarchical semantic reasoning to appearance-robust and physically consistent functional region grounding. 
As shown in Fig.~\ref{pipline}, the framework consists of three core modules: First, the Tree-structured Affordance Chain-of-Thought (TA-CoT) decomposes high-level task intentions into affordance-semantic instructions (see Sec.~\ref{tree-like}); second, a Spatially-Constrained Boundary Refinement (SCBR) loss is introduced to suppress boundary leakage and enhance spatial consistency (see Sec.~\ref{loss}); and finally, the Interactive  Convergent Refinement Flow (ICRF) is designed to aggregate dispersed affordance responses, achieving fine-grained manipulation point modeling that is physically consistent (see Sec.~\ref{sec:interactive_convergent_refinement_flow_field}).

\begin{figure}[t!]
\centerline{\includegraphics[width=0.48\textwidth]{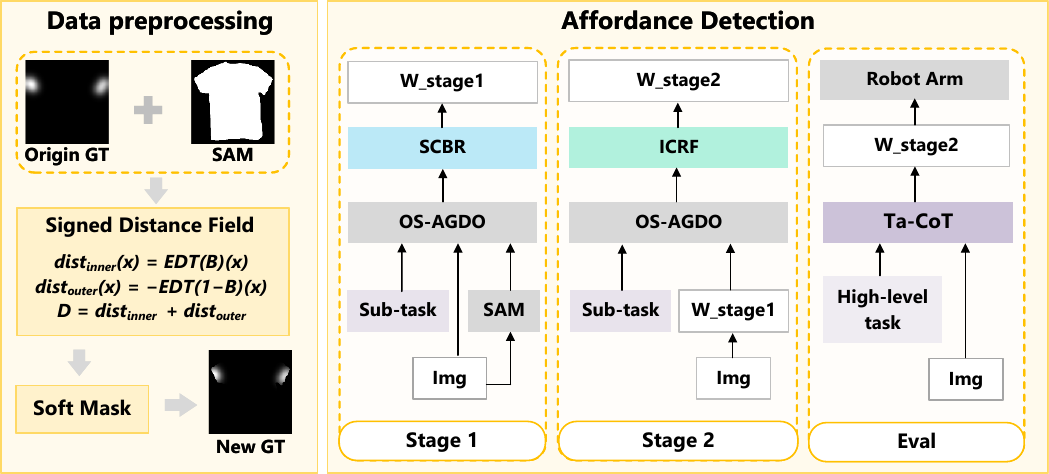}}
\captionsetup{font=small}
    \caption{\small Overview of the Fine-AGDDO15 dataset construction pipeline, together with the two-stage training and evaluation strategy.}
\label{data}
\vskip-3ex
\end{figure}

The proposed framework builds upon the OS-AGDO architecture~\cite{jia2025one}. As illustrated in Fig.~\ref{data}, the overall training process consists of two stages.

In Stage 1, we input the raw image mask processed by the Segment Anything Model (SAM)~\cite{Kirillov_2023_ICCV} and text instructions into an OS-AGDO backbone. To focus on interaction points, we remove the edge-extracting OEKRM module and apply SCBR loss for spatial constraints, yielding weights $W_{stage1}$.
In Stage 2, input the raw image and sub-task instruction, load $W_{stage1}$, and integrate the ICRF module to optimize for spatially continuous grounding regions, resulting in the final weights $W_{stage2}$.
During inference, the system leverages the dual-stage optimized weights and introduces the TA-CoT module to decouple high-level reasoning from perceptual localization. This separation isolates semantic uncertainties from grounding performance, enabling an objective evaluation of the mechanism’s physical robustness.

\subsection{Tree-structured Affordance Chain-of-Thought}
\label{tree-like}
In long-horizon manipulation of deformable objects, mapping abstract instructions to sub-task affordance semantics faces a significant semantic gap. To bridge this, we propose TA-CoT, a hierarchical reasoning mechanism using a Tree-structured Affordance Chain-of-Thought.
This mechanism decomposes the overall manipulation task into multiple interdependent sub-affordance tasks, transforming the reasoning process from a single-step decision into a hierarchically unfolded, temporally explicit, and progressive decision-making process.

\textbf{Formal Modeling of the Semantic Decision Tree.} 
We model affordance reasoning as a traversal process on a semantic decision tree $\mathcal{T}$. Given an egocentric visual observation $P_{img}$ and a language instruction $L$, our objective is to generate a sequence of affordance-semantic instructions $\mathcal{S}_{aff}=\{s^{aff}_1,s^{aff}_2,\dots,s^{aff}_N\}$.

First, the system determines the root node category $O$, partitioning the object space into two mutually exclusive sets:
%\vspace{-1.2ex}
\begin{equation}
% \[
O \in \mathcal{O}_{\text{Single}} \cup \mathcal{O}_{\text{Multi-step}}
% \]
\end{equation}
where $\mathcal{O}_{\text{Multi-step}}$ represents deformable objects requiring multi-step hierarchical affordance reasoning (\textit{e.g.}, clothes, pants), while $\mathcal{O}_{\text{Single}}$ denotes objects requiring only a single affordance-semantic instruction (\textit{e.g.}, hat, tissue).
Based on the active tree branch, TA-CoT decomposes the semantic reasoning process into a series of attribute-conditioned affordance selections:
\begin{equation}
P(\mathcal{S}_{aff} \mid O) = \prod_{k=1}^{K}
P(s^{aff}_k \mid O,c_k),
\end{equation}
where $s^{aff}_k$ denotes the affordance instruction selected at the $k$-th tree layer, and $c_k$ represents the corresponding structural or morphological attribute (\textit{e.g.}, $c_{\text{hood}}$) identified from the current visual state.
Sub-task progression is maintained by tree traversal and visual-state feedback, whereby the verified attribute $c_k$ determines the active branch, while inconsistent branches are rejected or kept dormant.

\textbf{TA-CoT Hierarchical Reasoning Mechanism.} 
TA-CoT organizes high-level deformable-object instructions through four progressive semantic levels:

(1)	Type Layer: category-level task stratification

This node serves as the entry point for reasoning, determining the task hierarchy according to the object category $O$ and whether deeper tree traversal is required. When $O \in \mathcal{O}_{\text{Single}}$, the output is a single-step affordance-semantic instruction $\mathcal{S}_{aff}$, where different categories correspond to specific textual affordance instructions (\textit{e.g.}, ``pull out'' for tissues, ``pull'' for curtains, ``hang'' for masks). 
For deformable objects in $\mathcal{O}_{\text{Multi-step}}$, the hierarchical inference tree is activated, and the system proceeds to the next semantic level.

(2)	Structure Layer: topological accessory perception and canonicalization

At this layer, the model focuses on the object's structural topology. Unlike ordinary surface wrinkles, accessory parts like hoods constitute structural topological protrusions that violate the quasi-planar geometric assumption required for garment folding. Therefore, if a hood is detected ($c_{\text{hood}}=1$), the model prioritizes the corresponding affordance sub-sequence $\mathcal{S}_{\text{hood}}$ to guide subsequent grounding for folding it back to its original position. This eliminates topological interference and canonicalizes the object's appearance into a standard rectangular topology, establishing a stable geometric foundation for the subsequent folding process. Conversely, for objects like pants or towels that lack topological accessories, this layer directly initializes the symmetry-based instruction sequence $\mathcal{S}_{\text{init}} = (pick \to place)$, avoiding unnecessary structural judgment and path expansion.

(3)	Attribute Layer: fine-grained morphology adaptation

This layer addresses morphological differences in the main body (\textit{e.g.}, sleeve length, pant leg length) and makes conditional decisions for morphology-related affordance-semantic instructions based on the current visual state. 
We define the morphology attribute set as $\mathcal{C}_{attr} = \{c_{\text{sleeve}}, c_{\text{leg}}\}$, where attributes are not predicted all at once but are confirmed progressively during the reasoning process by the multimodal model.
Unlike relying solely on static attributes, the model assesses the current spatial state of key parts before generating the next affordance-semantic instruction to determine whether the target specification has been met. Specifically, for sleeveless ($c_{\text{sleeve}}=\text{sleeveless}$) and short-sleeved ($c_{\text{sleeve}}=\text{short}$) garments, the model verifies whether the straps or cuffs have converged to the center region; for long-sleeved ($c_{\text{sleeve}}=\text{long}$) garments, it verifies whether the cuffs are aligned with the bottom hem. If state feedback indicates that the current configuration does not meet the target specification, the corresponding semantic instruction (\textit{e.g.}, $grasp\_sleeve \to put\_center/hem$) is triggered. This prevents redundant semantic steps and maintains reasoning consistency. For pants ($c_{\text{leg}}=\text{long}$), this layer appends a secondary folding instruction for the pant legs to complete the overall morphological canonicalization.

(4)	Finalization Layer: final shape canonicalization

The final layer selects the folding instruction for shape canonicalization, ensuring that the object ultimately converges to a standardized folded state. For upper-body garments and dresses, we adopt a unified shoulder-to-hem folding instruction, formalized as ($grasp\_shoulder \rightarrow put\_hem$).

\begin{figure}[t!]
\centerline{\includegraphics[width=0.48\textwidth]{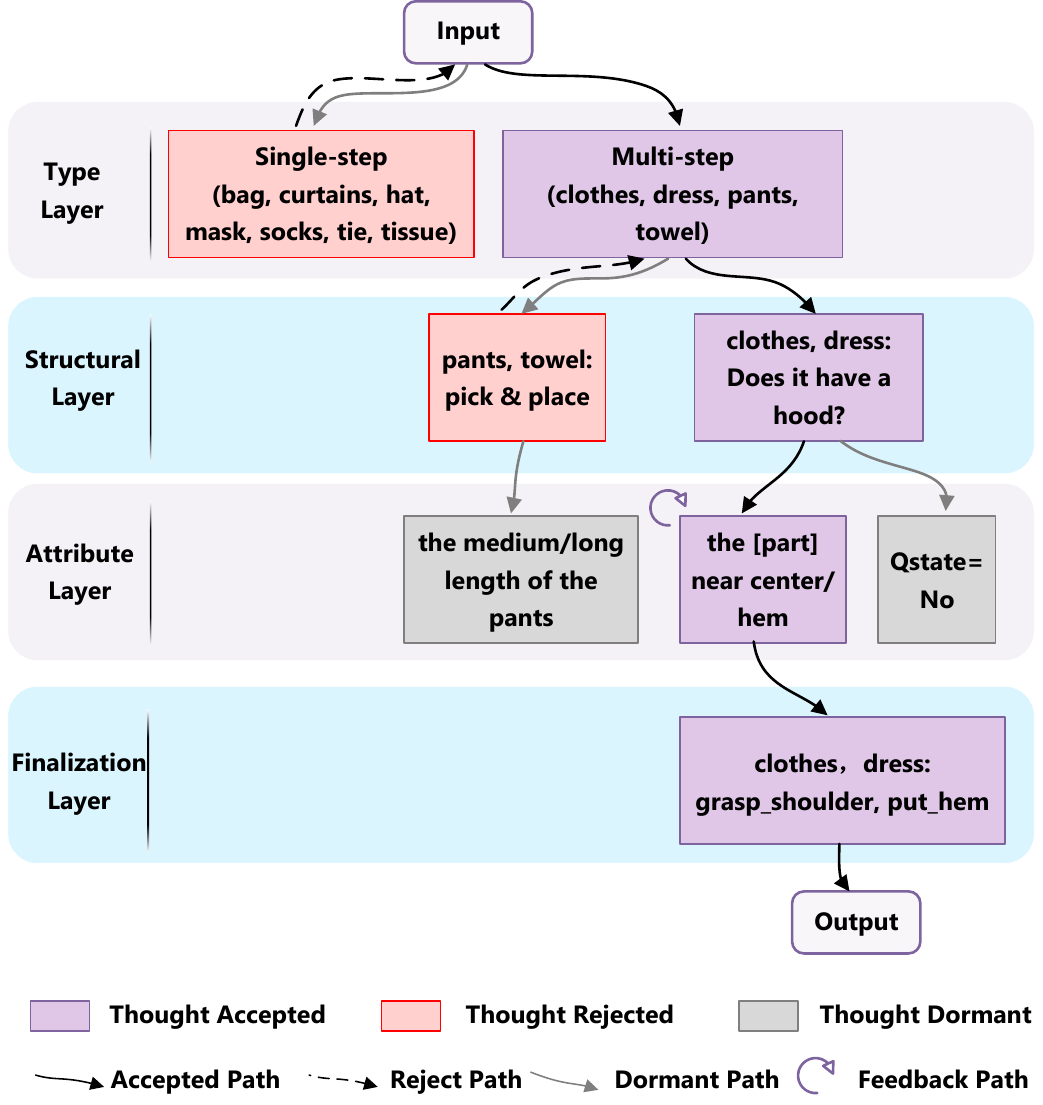}}
\captionsetup{font=small}
\caption{\small Four-state gating mechanism for TA-CoT reasoning. Hierarchical paths are dynamically managed via accept, reject, dormant, and feedback states.}
\label{cot}
\vskip-2ex
\end{figure}

\textbf{State Perception and Closed-Loop Routing Mechanism in the Structural Layer.}
To improve reasoning efficiency and avoid redundant branches, TA-CoT adopts a state-based routing mechanism with an accept–reject–dormant gating strategy to dynamically control the tree-structured reasoning path. 
As illustrated in Fig.~\ref{cot}, taking the clothes-with-hood category as an example: branches consistent with the current visual and semantic state are activated (Accept), while mismatched branches (\textit{e.g.}, pants or towels) are blocked (Reject), and their downstream nodes automatically become Dormant. In the Attribute Layer, a Feedback path further verifies visual states so that the corresponding affordance-semantic instruction is activated only when key parts have not reached their target configurations; otherwise, the branch is skipped.
This state-aware routing mechanism prunes invalid branches while maintaining structural integrity. By filtering redundant computation and updating the active reasoning path according to visual-state feedback, TA-CoT yields a compact, efficient, and physically consistent decision process.

Through this state perception and closed-loop routing mechanism, TA-CoT reduces invalid reasoning branches while preserving structural consistency, making the overall decision process more compact, efficient, and better aligned with the object's real-world physical constraints.

\subsection{Spatially-Constrained Boundary Refinement}
\label{loss}
\begin{figure}[t!]
\centerline{\includegraphics[width=0.44\textwidth]{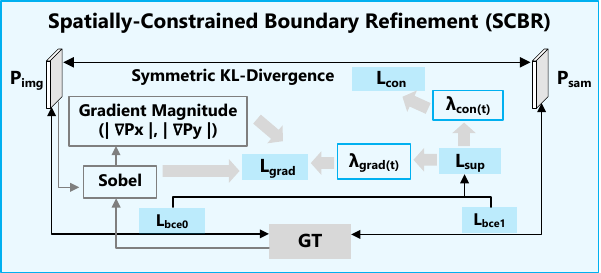}}
\captionsetup{font=small}
\caption{\small Illustration of SCBR, which suppresses boundary spillover and improves the spatial integrity of affordance predictions.}
\label{SCBR}
\vskip-2ex
\end{figure}

To constrain affordance predictions and prevent overflow, we formulate the Spatially-Constrained Boundary Refinement (SCBR) loss, integrating dual-stream supervision, structure-aware consistency, and boundary gradient penalties (Fig.~\ref{SCBR}). This strategy explicitly guides predictions to converge within physical boundaries, ensuring smooth and precise transitions. The total loss $\mathcal{L}_{total}$ is defined as:
\begin{equation}
% \[
\mathcal{L}_{total} = \mathcal{L}_{sup} + \lambda_{con}(t) \cdot \mathcal{L}_{con} + \lambda_{grad}(t) \cdot \mathcal{L}_{grad},
% \]
\end{equation}
where $\lambda_{con}(t)$ and $\lambda_{grad}(t)$ are weight coefficients dynamically adjusted based on the dual-stream supervision loss.

\textbf{Dual-Stream Supervision Loss ($\mathcal{L}_{sup}$).} 
The model comprises two synergistic prediction branches: an image-based branch $P_{img}$ derived from raw visual features, and a semantic-enhanced branch $P_{sem}$ integrated with SAM-generated masks. 
To ensure both independently converge to the true affordance region, we employ Binary Cross-Entropy (BCE) as the base supervision loss:
%\vspace{-1.2ex}
\begin{equation}
% \[
\mathcal{L}_{sup} = {L}_{\text{BCE}}(P_{img}, G) + {L}_{\text{BCE}}(P_{sem}, G),
% \]
\end{equation}
where $G$ is the normalized ground truth heatmap. This loss constrains the predicted regions to cover the target object.

\textbf{Structure-Aware Consistency Loss ($\mathcal{L}_{con}$).} 
The surfaces of deformable objects often exhibit complex textures (\textit{e.g.}, wrinkles, prints), causing the vision-only branch $P_{img}$ to be highly vulnerable to background and textural noise. Conversely, while the mask-guided branch $P_{sem}$ suppresses background interference, it may overfit to non-affordance regions. To leverage their synergistic strengths, we introduce a Kullback-Leibler (KL) divergence to induce semantic convergence:
%\vspace{-1.2ex}
\begin{equation}
% \[
\mathcal{L}_{con} = \frac{1}{2} \left( D_{KL}(P_{img} || P_{sem}) + D_{KL}(P_{sem} || P_{img}) \right).
% \]
\end{equation}
This loss promotes alignment at the semantic level, effectively filtering out background interference, suppressing texture-induced overfitting, and enhancing prediction robustness.

\textbf{Boundary Gradient Penalty Loss ($\mathcal{L}_{grad}$).} 
To constrain the gradient of the predicted map at object boundaries and promote smooth inward convergence of the activated region, we design a gradient penalty loss based on the Sobel operator.
%\vspace{-1.2ex}
\begin{equation}
% \[
\mathcal{L}_{grad} = \mathbb{E} \left[ (\mid \nabla P_x \mid \cdot \mathcal{M}_{bound})^2 + (\mid \nabla P_y \mid \cdot \mathcal{M}_{bound})^2 \right],
% \]
\end{equation}
where $\nabla_x$ and $\nabla_y$ denote the horizontal and vertical gradients of the predicted affordance map, respectively, and $\mathcal{M}_{bound}$ is the boundary mask calculated from the Ground Truth (GT) boundary. 
By minimizing the gradient at boundaries, this loss suppresses spurious activations and abrupt fluctuations, encouraging smooth transitions that align with the object’s physical contours.

\textbf{Dynamic Weight Balancing Mechanism.} 
To coordinate the learning focus across training stages, we adopt a dynamic weighting strategy, where the coefficient $\lambda(t)$ is inversely proportional to the dual-stream supervision loss $\mathcal{L}_{sup}$:
%\vspace{-1.2ex}
\begin{equation}
% \[
\lambda(t) \propto \frac{1}{\mathcal{L}_{sup}(t) + \epsilon}.
% \]
\end{equation}
In the initial training phase, the model is dominated by $\mathcal{L}{sup}$ and focuses on coarse-grained learning of global affordance regions. As training progresses, the weights of $\mathcal{L}{con}$ and $\mathcal{L}_{grad}$ gradually increase, guiding the model toward finer boundary alignment and structural consistency optimization, thereby achieving progressive coarse-to-fine boundary refinement.

\subsection{Interactive Convergent Refinement Flow}
\label{sec:interactive_convergent_refinement_flow_field}
Although the first stage preliminarily corrects object boundaries through spatial constraints, the initial affordance distribution ($x_0$) often remains topologically fragmented and locally discontinuous.  
This primarily stems from misalignment between low-level visual features and high-level semantic reasoning, as well as uncertainty introduced by complex texture patterns. 
To address this core challenge, we establish an Interactive Convergent Refinement Flow (ICRF). 
Starting from the first-stage affordance map as the initial state $x_0$, ICRF simulates a convergence process driven by adaptive acceleration. This progressively refines scattered affordance responses toward a spatially coherent and semantically explicit target distribution $x_1$.

\begin{algorithm}[t]
    \newcommand\mycommfont[1]{\footnotesize\ttfamily\textcolor{gray}{#1}}
    \SetCommentSty{mycommfont}
    \SetKwComment{Comment}{//\ }{}
    \caption{ICRF Acceleration Learning and Dynamic Refinement}
    \label{alg:icrf}

    \KwIn{Stage-1 affordance map $x_0$, target map $x_1$,
    velocity steps $N_\tau$, state steps $N_t$}
    \KwOut{Learned acceleration field $a_\theta$ and refined map $x_{\mathrm{final}}$}

    \Comment{Acceleration Field Learning}
    \While{not converged}{
        Sample $t,\tau \sim \mathcal{U}[0,1]$\;

        $x_t \leftarrow (1-t)x_0+t x_1,\quad
        v_0 \leftarrow x_1-x_t$\;

        $v_{\mathrm{target}} \leftarrow x_1-x_0,\quad
        v_\tau \leftarrow (1-\tau)v_0+\tau v_{\mathrm{target}}$\;

        $a_{gt} \leftarrow v_{\mathrm{target}}-v_0$\;

        $\mathcal{L}_{flow}\leftarrow
        \|a_\theta(v_\tau,\tau,x_t,t)-a_{gt}\|^2$\;

        Update $\theta$ via $\nabla_\theta\mathcal{L}_{flow}$\;
    }

    \Comment{Dynamic Refinement}
    $x \leftarrow x_0,\quad
    v \sim \mathcal{N}(0,0.1^2 I)$\;

    Define $\Delta t=1/N_t,\quad
    \Delta\tau=1/N_\tau$\;

    \For{$i\leftarrow0$ \KwTo $N_t-1$}{
        $t_i\leftarrow i\Delta t$\;

        \For{$j\leftarrow0$ \KwTo $N_\tau-1$}{
            $\tau_j\leftarrow j\Delta\tau$\;
            $a_{\rm pred}\leftarrow
            a_\theta(v,\tau_j,x,t_i)$\;
            $v\leftarrow v+a_{\rm pred}\Delta\tau$\;
        }

        $x\leftarrow x+v\Delta t$\;
    }

    \KwRet{$a_\theta,\;x_{\mathrm{final}}\leftarrow x$}
\end{algorithm}

\begin{figure}[t!]
\centerline{\includegraphics[width=0.48\textwidth]{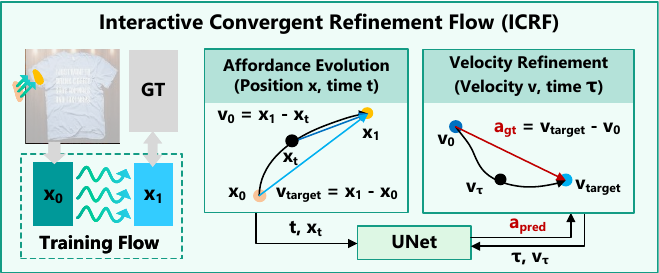}}
\captionsetup{font=small}
\caption{\small Illustration of ICRF, which progressively aggregates dispersed responses into compact and spatially coherent affordance regions.}
\label{ICRF}
\vskip-2ex
\end{figure}

\textbf{Adaptive Acceleration Flow Modeling.} 
Existing flow matching frameworks typically model state transitions through a first-order velocity field $v$. However, a first-order velocity field directly models state evolution but does not explicitly characterize how the refinement velocity itself should be adjusted for spatially heterogeneous affordance responses. To address this limitation, we introduce a second-order refinement process driven by a learnable acceleration field $a_\theta$, which explicitly models the progressive correction of the refinement velocity.
Specifically, during acceleration-field learning, \(x_1\) denotes the ground-truth target affordance distribution. We define two coupled flow processes to describe this dynamic evolution (Fig.~\ref{ICRF}). The first is the affordance state flow on time $t \in [0, 1]$, describing the evolution of the affordance-response map from the initial distribution $x_0$ to the target $x_1$:
\begin{equation}
% \[
x_t = (1-t)x_0 + t x_1.
% \]
\end{equation}
The second component introduces an auxiliary temporal axis $\tau \in [0, 1]$ in the velocity space to model the progressive evolution of the refinement velocity, constructing a velocity flow from the initial residual velocity $v_0=x_1-x_t$ to the target velocity $v_{target} =x_1-x_0$:
\begin{equation}
% \[
v_\tau = (1-\tau)v_0 + \tau (x_1 - x_0).
% \]
\end{equation}
From this velocity evolution, we derive the target acceleration field that governs the refinement dynamics as:
%\vspace{-1.2ex}
\begin{equation}
% \[
a_{gt} = \frac{\partial v_\tau}{\partial \tau} = (x_1 - x_0) - v_0.
% \]
\end{equation}
The objective is to train a parameterized acceleration field $a_\theta$ that approximates $a_{gt}$. By minimizing $\mathcal{L}_{flow} = \| a_\theta - a_{gt} \|^2$, the model learns the progressive velocity correction required for subsequent refinement.

\textbf{Dynamic Convergent Inference.} 
During inference, the Stage-1 affordance map is initialized as $x_0$, while the refinement velocity is initialized with small Gaussian noise. The learned acceleration field $a_\theta$ then drives the following coupled ODEs:
%\vspace{-1.2ex}
\begin{equation}
% \[
\begin{cases} \frac{d v_\tau}{d \tau} = a_\theta(v_\tau, \tau, x_t, t), & \tau \in [0, 1] \quad \text{(Velocity Refinement)} \\ \frac{d x_t}{d t} = v_{\tau=1}(x_t, t). & t \in [0, 1] \quad \text{(Affordance Evolution)} \end{cases}
% \]
\end{equation}
At each state step $t$, the learned acceleration field $a_\theta$ is integrated along the auxiliary temporal axis $\tau$ to progressively refine the velocity field. The resulting velocity $v_{\tau=1}$ is then used to update the affordance state, and this alternating process is repeated until the final refined map is obtained. Through this acceleration-driven refinement, dispersed responses remaining after the first stage are consolidated into compact and spatially coherent affordance regions.

\section{Experiments}

\subsection{Experiment Setup}

\textbf{Datasets.} 
To enable precise manipulation and robust affordance grounding, we present Fine-AGDDO15, an enhanced version of AGDDO15~\cite{jia2025one}. While AGDDO15 provides egocentric, one-shot affordance annotations for $15$ object classes and $15$ affordance types, it mainly supports coarse semantic grounding, leaving pixel-level boundaries under-optimized for high-precision tasks. We retain the original taxonomy and apply three key refinements (Fig.~\ref{data}):
(1) Type completion. We conduct a comprehensive supplementation and correction of affordance interaction types that were omitted for certain objects in the original dataset. 
(2) Boundary constraint and background removal. 
To rigorously eliminate erroneous annotations that spilled into background regions, we leverage SAM~\cite{Kirillov_2023_ICCV} to generate high-precision object masks and perform an intersection operation with the original GT.
(3) Soft mask generation. 
To address the information loss inherent in binary masks at boundaries, we introduce a soft mask technique based on a composite distance field. This technique fuses the Euclidean distance of foreground pixels ($dist_{inner}$) with the negative distance of background pixels ($dist_{outer}$) to construct a continuous field. 
A sigmoid function is then employed to map discrete binary signals into soft masks with smooth transitions within the $[0,1]$ interval. 
This improves edge detail and boundary supervision during training.

\textbf{Implementation Details.} 
Experiments are conducted on a single NVIDIA RTX 3090 GPU. 
All input images were resized and cropped to a resolution of $224{\times}224$. 
The training consists of two stages: 
(1) In the first stage, the model is trained using the SGD optimizer with an initial learning rate of $1{\times}10^{-2}$ for $20,000$ iterations. 
(2) In the second stage, the acceleration field network is optimized using the Adam optimizer with a learning rate of $2{\times}10^{-7}$. 
During this stage, we employ a learning rate scheduling strategy consisting of linear warmup followed by cosine annealing, training for a total of $50,000$ iterations.

\begin{figure*}[t!]
\centerline{\includegraphics[width=1\textwidth]{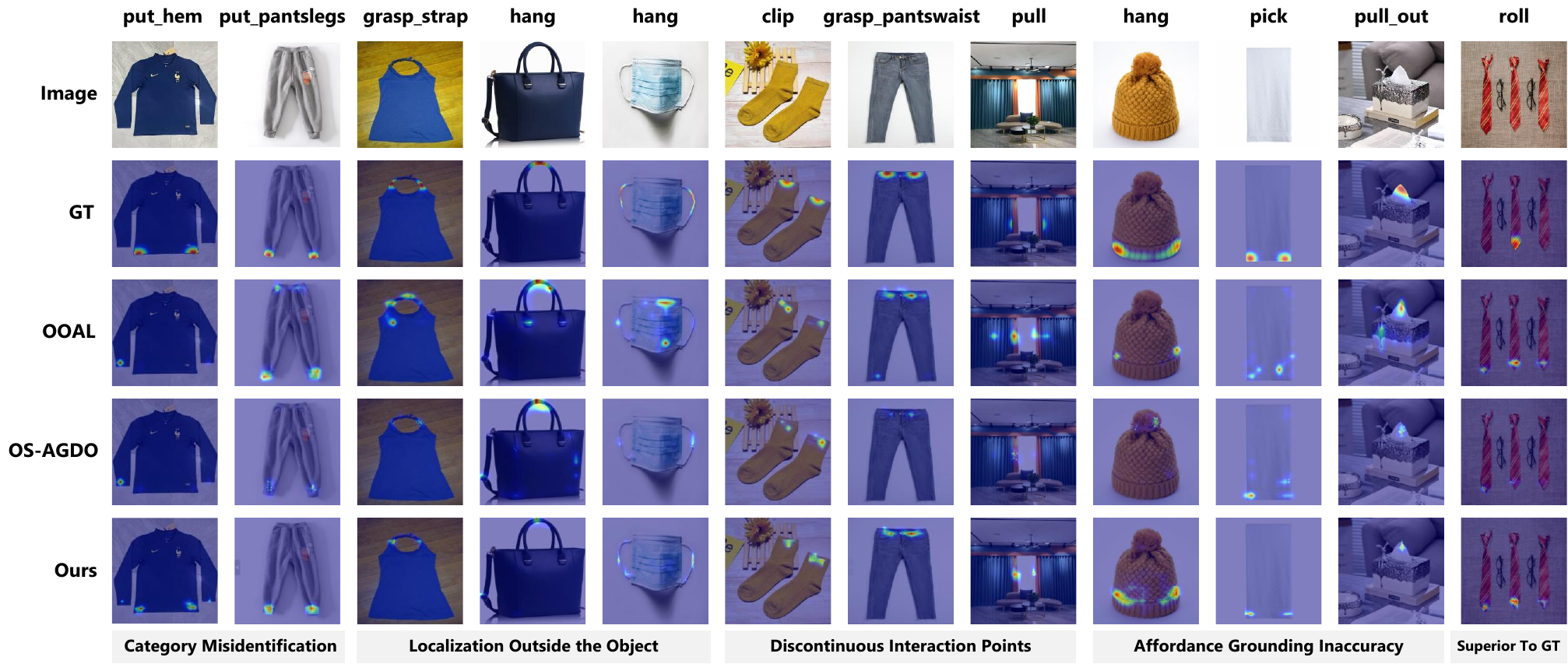}}
\captionsetup{font=small}
\caption{\small Qualitative comparison of affordance grounding results. We compare our method with two one-shot OSTA baselines (OOAL~\cite{li2024one} and OS-AGDO~\cite{jia2025one}) and the GT. 
Our approach consistently outperforms the baselines, yielding affordance regions that are more spatially accurate, structurally coherent, and physically plausible, closely aligning with the GT.
}
\label{viz}
\vskip-2ex
\end{figure*}

\textbf{Metrics.} 
We conduct evaluations from two perspectives:
(1) Affordance Grounding: Three standard metrics are used, specifically Kullback-Leibler Divergence (KLD) for distribution difference, Similarity (SIM) for histogram overlap, and Normalized Scanpath Saliency (NSS) for response intensity at ground-truth fixation points.
(2) Real-World Manipulation: Success rate (SR) measures task performance. For folding, the system plans instructions via CoT and generates affordance regions to guide ``Pick-and-Place'' actions. For pull-out, it identifies the optimal grasp point and transfers the object to a target zone. For hanging, the mask or hat is placed stably on the rack. Success requires successful task completion without dropping or placement failure.

\begin{table}[!t]
    \centering
    \renewcommand{\arraystretch}{1.2}
    \caption{Comparison of different MLLMs as the reasoning core of TA-CoT.}
    \label{cot_com}
    \setlength{\tabcolsep}{3.5mm}
    \begin{tabular}{l c}
        \hline
        \textbf{Model} & \textbf{Accuracy} ($\uparrow$) \\
        \hline\hline
        Claude 3.7         & 82.15\% \\
        Gemini-2.5-Flash   & 85.14\% \\
        DeepSeek-Chat      & 74.21\% \\
        GPT-4o             & 80.87\% \\
        InternVL2-4B       & \textbf{90.61\%} \\
        \hline
    \end{tabular}
    \vskip-3ex
\end{table}

\subsection{Comparative Evaluation of Chain-of-Thought Models}
As illustrated in Table~\ref{cot_com}, we present the comparative experimental results of utilizing different Multimodal Large Language Models (MLLMs) as the semantic reasoning core within the TA-CoT module. 
The results indicate that InternVL2-4B~\cite{chen2024internvl} achieves the highest overall accuracy ($90.61\%$) and demonstrates superior decision-making robustness in hierarchical affordance reasoning tasks for deformable objects. 
In contrast, although GPT-4o~\cite{hurst2024gpt}, Gemini-2.5-flash~\cite{comanici2025gemini}, and Claude 3.7 possess powerful general-purpose language understanding capabilities, they exhibit semantic drift and decision instability when handling tree-structured reasoning that involves multi-level conditional dependencies and fine-grained structural attribute decomposition. Meanwhile, Deepseek-chat ($74.21\%$), serving as a text-only baseline, significantly lags in performance due to its inability to perceive visual information, forcing it to rely solely on text instructions for abstract Chain-of-Thought reasoning. 
Given the complexity of fine-grained morphological alignment for deformable objects and the requirements of robotic manipulation for efficient reasoning, we ultimately selected InternVL2-4B as the core reasoning engine to balance accuracy and deployment efficiency.

\subsection{Results of Affordance Grounding}
We evaluate the efficacy of TRACER on the Fine-AGDDO15 dataset through comprehensive qualitative and quantitative analyses. 
Unlike VLA-based paradigms requiring extensive action data for fine-tuning, our framework enables lower-cost deployment through one-shot reasoning.
To address the scarcity of research in egocentric one-shot affordance grounding, we compare against four representative methods: RAGNet~\cite{wu2025ragnet}, which performs reasoning-based affordance segmentation, Affordance-R1~\cite{wang2025affordancer1}, which enhances generalizable affordance reasoning with multimodal reinforcement learning, OOAL~\cite{li2024one}, a vision-language transfer framework, and OS-AGDO~\cite{jia2025one}, which models local structural and geometric cues for deformable objects. These comparisons evaluate our framework under complex texture interference and fine-grained spatial grounding requirements.

\textbf{Quantitative Results.} 
As shown in Table~\ref{pre_table}, our method achieves the best overall grounding performance, improving KLD, SIM, and NSS over OS-AGDO by $4.8\%$, $7.5\%$, and $4.3\%$, respectively. RAGNet and Affordance-R1, despite their strong reasoning capabilities, are less adapted to one-shot egocentric grounding of deformable objects, leading to less concentrated and spatially consistent predictions. OOAL benefits from vision-language transfer, while OS-AGDO improves structural modeling, but both lack explicit spatial refinement for fine-grained interaction regions. In contrast, SCBR and ICRF jointly suppress boundary overflow and aggregate dispersed responses, producing more compact and coherent affordance regions. Regarding efficiency, RAGNet and Affordance-R1 operate below $1$ FPS, whereas OOAL and OS-AGDO achieve $50.23$ and $20.35$ FPS. Our method reaches $6.54$ FPS, offering a favorable balance between grounding accuracy and computational efficiency.

\begin{table}[!t]
\centering
\captionsetup{font=small}
\renewcommand{\arraystretch}{1.3}
\setlength{\tabcolsep}{2.6pt}
\scriptsize
\caption{\small Comparison with state-of-the-art methods on the Fine-AGDDO15 dataset in terms of grounding accuracy and computational efficiency. The \textbf{best} grounding results are highlighted in bold. ($\uparrow$/$\downarrow$ indicates higher/lower is better).}
\label{pre_table}
\resizebox{\columnwidth}{!}{
\begin{tabular}{@{}l|ccc|ccc@{}}
\hline
\textbf{Model} & \textbf{KLD} ($\downarrow$) & \textbf{SIM} ($\uparrow$) & \textbf{NSS} ($\uparrow$) & \textbf{Params} ($\downarrow$) & \textbf{FLOPs} ($\downarrow$) & \textbf{FPS} ($\uparrow$) \\
\hline\hline
RAGNet~\cite{wu2025ragnet} & 5.752 & 0.122 & 0.709 & 7704.76 & 51738.41 & 0.605 \\
Affordance-R1~\cite{wang2025affordancer1} & 5.388 & 0.239 & 1.694 & 8516.60 & 23400.44 & 0.130 \\
OOAL~\cite{li2024one} & 1.297 & 0.404 & 2.819 & 135.95 & 135.57 & 50.23 \\
OS-AGDO~\cite{jia2025one} & 1.257 & 0.412 & 2.866 & 136.46 & 135.81 & 20.35 \\
\textbf{Ours} & \textbf{1.197} & \textbf{0.443} & \textbf{2.990} & 219.00 & 1889.46 & 6.54 \\
\hline
\end{tabular}}
\vskip-2ex
\end{table}

\textbf{Qualitative Analysis.}
As shown in Fig.~\ref{viz}, we qualitatively compare affordance grounding results from two OSTA baselines (OOAL and OS-AGDO), our approach, and GT. 
Overall, our method consistently surpasses the baselines in spatial precision, structural coherence, and manipulation relevance, producing affordance regions that are more compact, continuous, and physically executable. Representative observations are summarized as follows:

(1) Category Misidentification (Cols. 1–2). While baseline methods frequently exhibit semantic confusion or misaligned activation zones, the proposed approach consistently identifies the correct category under semantic guidance and constrains predictions within plausible object boundaries via spatial refinement.
(2) Localization Outside the Object (Cols. 3–5). For slender objects such as ropes, baselines tend to generate coarse and overflowing responses due to weak spatial convergence. In contrast, our method produces highly concentrated predictions that closely adhere to the object’s topological and geometric structure.
(3) Discontinuous Interaction Points (Cols. 6–8). Baselines tend to produce fragmented activations, resulting in selection uncertainty. Our method effectively integrates multiple potential sub-regions into spatially connected and semantically clear representations, providing stable physical references for grasp planning.
(4) Affordance Grounding Inaccuracy (Cols. 9–11). Our framework exhibits stronger spatial discriminative power, significantly suppressing non-functional activations outside the target region and achieving higher localization precision.
(5)  Generalization Beyond Ground Truth (Col. 12). Despite the GT only labeling a single instance, our approach exhibits remarkable generalization, predicting structurally complete and physically plausible affordance regions on unlabeled objects of the same class, often surpassing the original annotations in terms of completeness.

\begin{table}[!t]
    \centering
    \renewcommand{\arraystretch}{1.3}
    \caption{Ablation study on loss components and weighting strategies. $\mathcal{L}_{sup}$, $\mathcal{L}_{con}$, and $\mathcal{L}_{grad}$ denote dual-stream supervision, structure-aware consistency, and boundary gradient penalty, respectively.}
    \label{tab:ablation_loss}
    \setlength{\tabcolsep}{2.5mm}
    \begin{tabular}{c c c c | c c c}
    \hline
    $\mathcal{L}_{sup}$ & $\mathcal{L}_{con}$ & $\mathcal{L}_{grad}$ & Strategy & KLD $\downarrow$ & SIM $\uparrow$ & NSS $\uparrow$ \\
    \hline\hline
    \checkmark & & & - & 1.264 & 0.434 & 2.805 \\
    \checkmark & \checkmark & & Dynamic & 1.248 & \textbf{0.437} & 2.832 \\
    \checkmark & & \checkmark & Dynamic & 1.239 & 0.432 & 2.857 \\
    \checkmark & \checkmark & \checkmark & Fixed & 1.229 & 0.432 & 2.903 \\
    \checkmark & \checkmark & \checkmark & Dynamic & \textbf{1.218} & 0.425 & \textbf{2.909} \\
    \hline
    \end{tabular}
    \vskip-2ex
\end{table}

\subsection{Ablation Studies.} 

\textbf{Analysis of Loss Components.}
Table~\ref{tab:ablation_loss} shows the impact of each spatial constraint and the dynamic weighting strategy. Results indicate that the full setup with dynamic weighting achieves the best performance (KLD: $1.218$, NSS: $2.909$), with each component showing distinct optimization effects:
(1) Structural Alignment.
Introducing structure-aware consistency ($\mathcal{L}_{con}$) alone significantly enhances shape similarity, boosting the SIM metric to a peak of $0.437$. This signifies a tighter alignment between the predicted distribution and the object’s inherent boundaries, thereby maximizing the spatial overlap with the ground truth.
(2) Boundary-Aware Contraction.
While the boundary gradient penalty ($\mathcal{L}_{grad}$) further optimizes KLD and NSS, it causes a slight decline in SIM. This suggests that stringent geometric constraints induce a contraction of predicted regions toward physical boundaries, prioritizing structural precision over area coverage.
(3) Dynamic Weighting Strategy.
We further examine the weighting mechanisms. 
Unlike the fixed strategy that employs constant coefficients, the dynamic strategy introduces an adaptive mechanism to balance gradient contributions by inversely scaling auxiliary loss weights relative to the primary supervision loss. This dynamic adjustment further reduces KLD to $1.218$ and elevates NSS to $2.909$, demonstrating superior convergence stability.

Notably, although the final model’s SIM ($0.425$) is slightly below the $\mathcal{L}{sup}+\mathcal{L}{con}$ peak, it remains comparable to the baseline. The $\mathcal{L}_{grad}$ term sharpens distributions by contracting heatmap boundaries toward physical contours, pruning low-confidence edge pixels. This is supported by the NSS improvement ($2.805 \rightarrow 2.909$), indicating a stronger focus on the affordance core and improved geometric certainty.

\begin{table}[!t]
    \centering
    \renewcommand{\arraystretch}{1.2}
    \caption{Impact of each design component on affordance grounding performance. Baseline*: OS-AGDO without the OEKRM module; SCBR: Spatially-Constrained Boundary Refinement; ICRF: Interactive Convergent Refinement Flow.}
    \label{tab:impact_design}
    \setlength{\tabcolsep}{4.5mm}
    \begin{tabular}{l | c c c}
    \hline
    Method & KLD $\downarrow$ & SIM $\uparrow$ & NSS $\uparrow$ \\
    \hline\hline
    Baseline* & 1.221 & 0.425 & 2.897 \\
    + SCBR (Loss) & 1.218 & 0.425 & 2.909 \\
    + SCBR + ICRF & \textbf{1.197} & \textbf{0.443} & \textbf{2.990} \\
    \hline
    \end{tabular}
    \vskip-2ex
\end{table}

\begin{figure}[!t]
  \centering
  \small
  \includegraphics[width=0.48\textwidth]{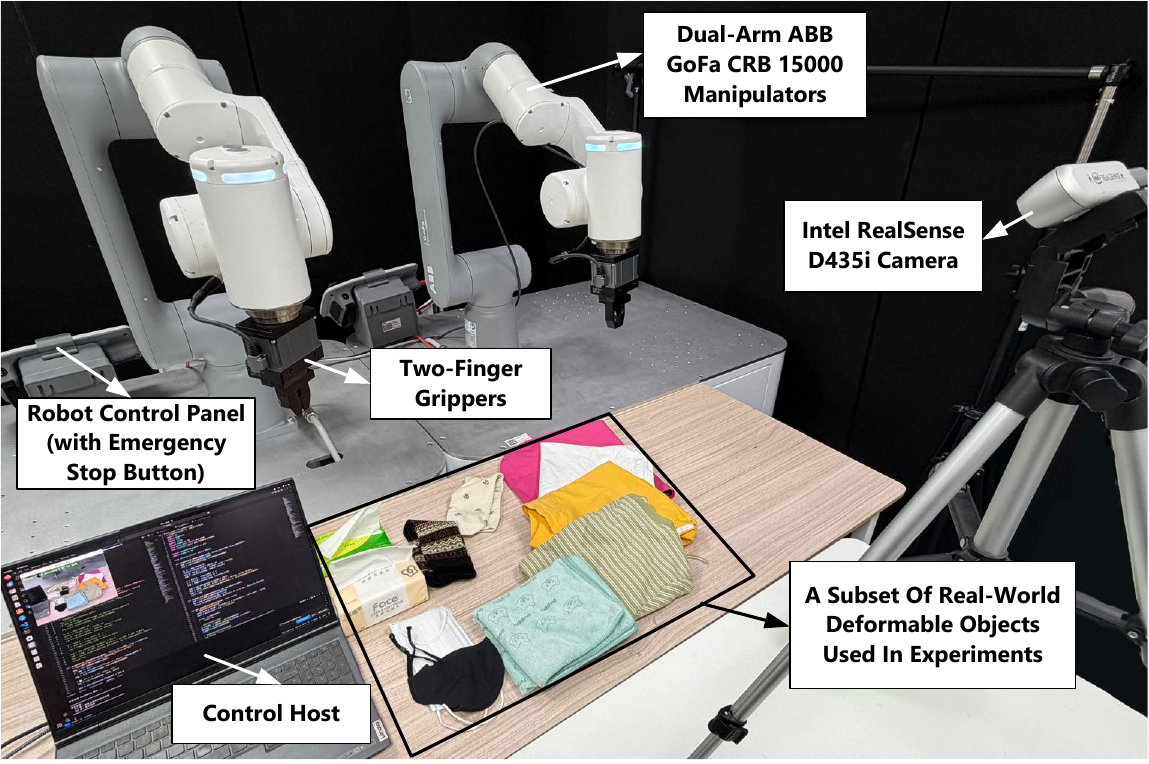}
  \captionsetup{font=small}
  \caption{\small 
  The system includes a dual-arm ABB GoFa CRB 15000 with two-finger grippers, an Intel RealSense D435i camera, and a control computer for model inference and task scheduling, supporting closed-loop deformable object manipulation.
  }
 \label{platform}
 \vskip-2ex
\end{figure}

\begin{figure*}[t!]
\centerline{\includegraphics[width=0.85\textwidth]{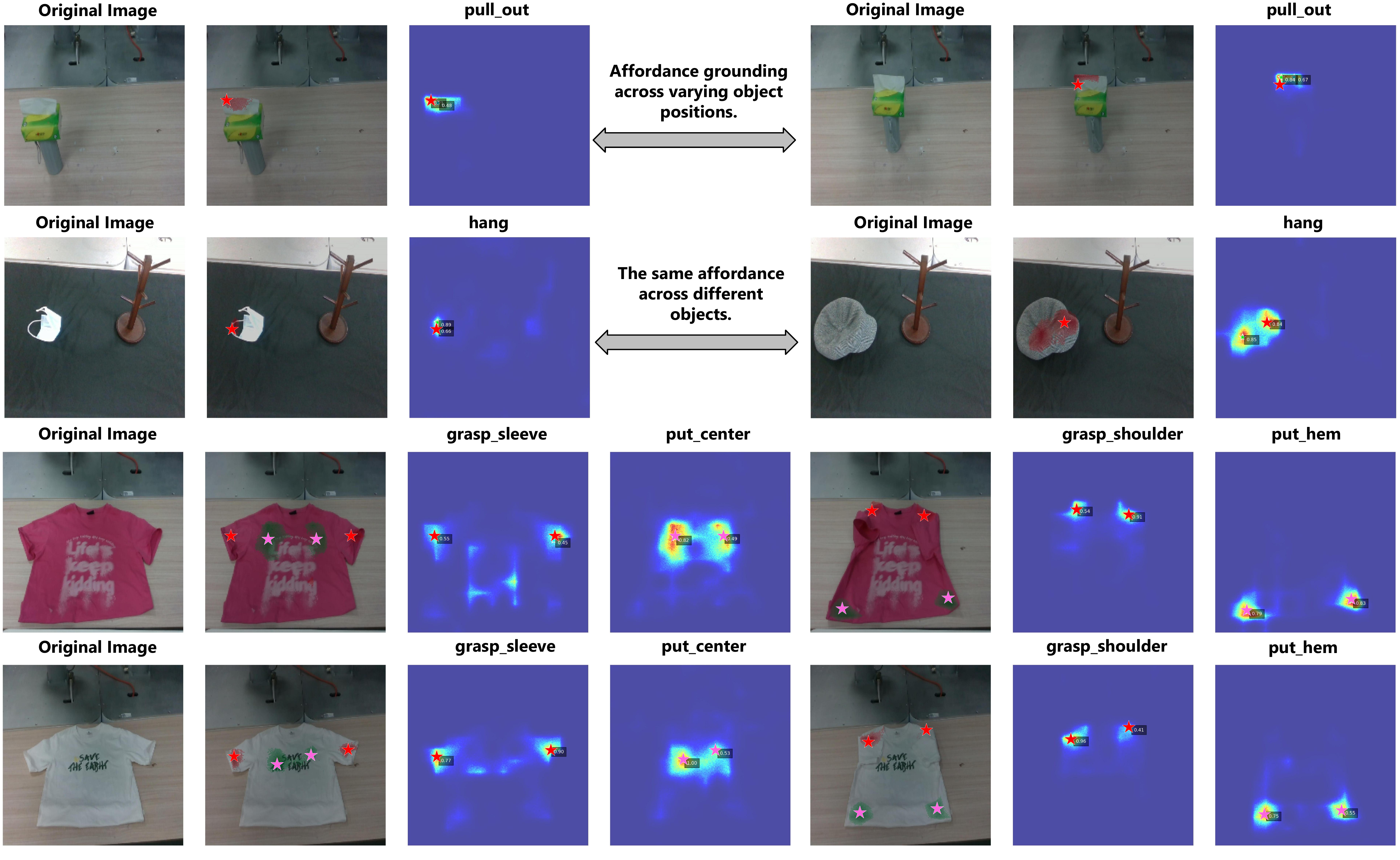}}
\captionsetup{font=small}
\caption{\small Qualitative analysis of affordance grounding in real-world experiments. The \textcolor{red}{$\star$} and \textcolor{black}{$\star$} pentagrams denote ``Pick'' and ``Place'' manipulation points, respectively, annotated on both prediction images and affordance heatmaps. The results demonstrate robust grounding under pose variations, across different object instances, and under diverse texture conditions.
}
\label{ex_process}
\vskip-2ex
\end{figure*}

\textbf{Impact of Each Design.} 
As shown in Table~\ref{tab:impact_design}, we verify the effectiveness of each module by progressively introducing core components.
(1) Effect of SCBR Loss.
In Stage 1, integrating SCBR loss into the OS-AGDO baseline yields consistent improvements in KLD and NSS. Although SIM remains at $0.425$, SCBR effectively rectifies coarse contours by concentrating the probability distribution on the object body, establishing a robust geometric prior for subsequent refinement.
(2) Effect of the ICRF Model.
In Stage 2, the ICRF module triggers a comprehensive performance leap: KLD drops to $1.197$, while NSS and SIM surge to $2.990$ and $0.443$. Building upon the global distribution shaped by SCBR, ICRF uses a dynamic acceleration field to aggregate discrete responses into a precise interaction center, achieving fine-grained sharpening and morphological alignment.
In summary, the SCBR module anchors the global probability distribution, while the ICRF module executes high-fidelity morphological refinement. Their synergistic interaction achieves the optimal performance across all evaluation dimensions.

\subsection{Performance in Real-World Scenarios}

\textbf{Experimental Setup and Deployment Details.} To thoroughly evaluate the deployment feasibility and practical performance of the proposed method in real-world scenarios, we established a real-world experimental platform, as illustrated in Fig.~\ref{platform}. The platform comprises a dual-arm 6-DoF ABB GoFa CRB 15000 collaborative robotic system, two ABB-compatible two-finger grippers, an Intel RealSense D435i depth camera, two robot control panels equipped with emergency stop functionality, and a variety of deformable objects. Additionally, the system is equipped with a control computer dedicated to model inference and task scheduling. The proposed model is deployed on this control host, where it receives language instructions and visual data captured by the camera. Through a local area network, the system communicates with the robot controller to perform task reasoning, affordance grounding, and execution of deformable object manipulation.

\textbf{Manipulation Execution Pipeline.} In the specific execution pipeline, the system first acquires visual data captured by the camera and employs the TA-CoT to infer the target affordance semantics for the current step. 
Subsequently, the trained two-stage model generates high-fidelity, spatially refined affordance heatmaps.
During the post-processing stage, clustering analysis is performed on these heatmaps to extract the optimal manipulation pixel coordinates for both the left and right manipulator arms. 
Finally, the predicted pixel coordinates are mapped to the robot base coordinate frame via the hand-eye calibration matrix. These translational components are then integrated with a predefined end-effector orientation to formulate the target SE(3) pose, which drives the dual-arm robot to execute synchronized manipulation tasks.

\begin{figure}[!t]
  \centering
  \small
  \includegraphics[width=0.48\textwidth]{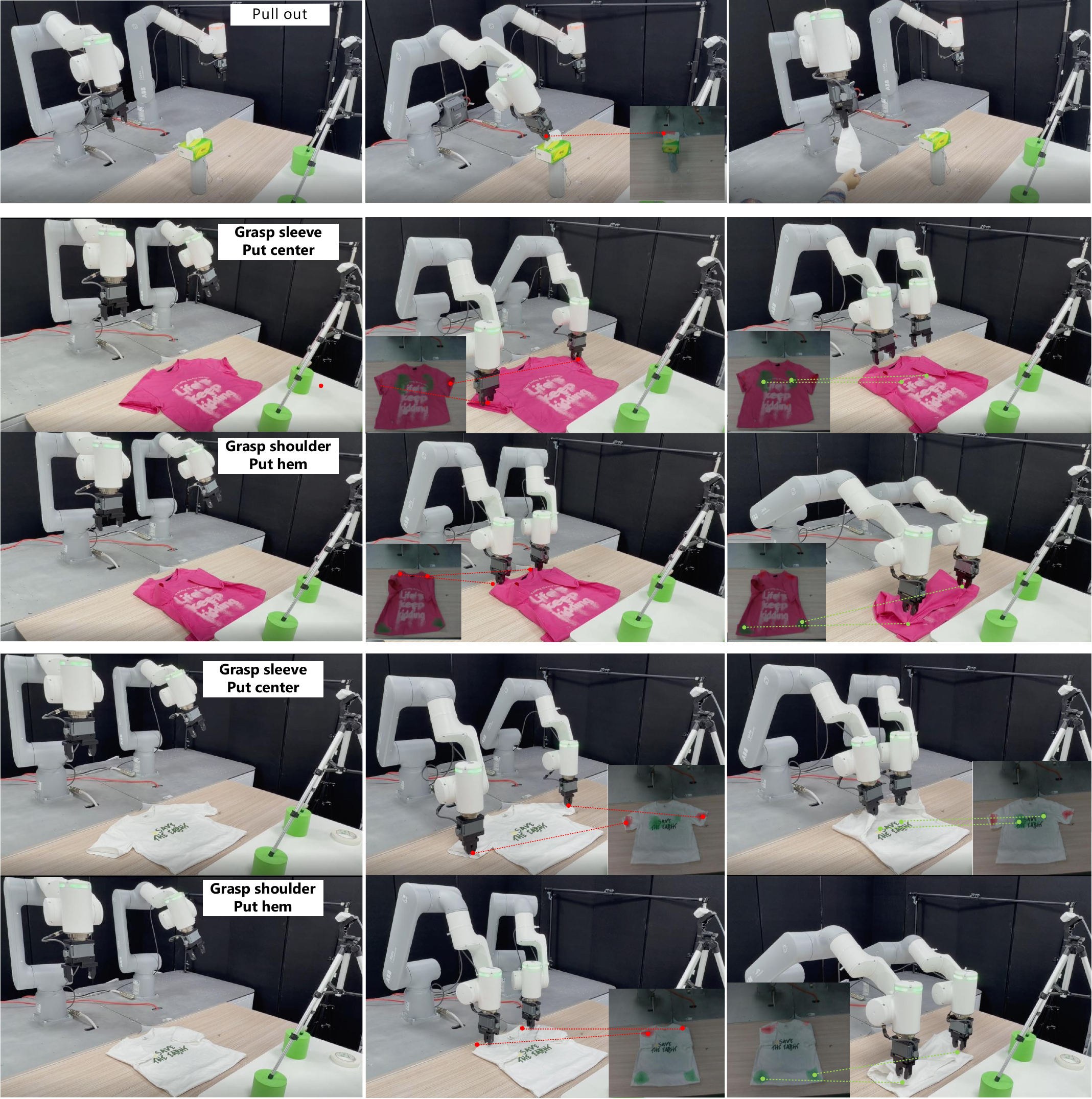}
  \captionsetup{font=small}
  \caption{\small Visualization of keyframe sequences and affordance grounding in real-world experiments.}
 \label{process}
 \vskip-1ex
\end{figure}

\textbf{Affordance Grounding in Representative Scenarios.} 
To qualitatively analyze the robustness of TRACER in complex environments, representative results are presented in Fig.~\ref{ex_process}. 
In the visualization, red pentagrams and black pentagrams denote the ``Pick'' and ``Place'' manipulation points, respectively. These points are consistently annotated on both the real-world affordance prediction image and the corresponding single-action heatmap. Key observations include:
(1) Pose Robustness. As shown in the first row, the framework reliably identifies physically executable points despite variations in object pose or configuration. For single-arm tasks, the system defaults to the right-arm output for stable execution.
(2) Cross-Instance Grounding. As shown in the second row, the same affordance can be consistently grounded across different object instances, demonstrating robust cross-instance generalization.
(3) Texture Robustness. As observed in the third and fourth rows, high-response regions remain concentrated in functional areas across diverse colors and textures, suppressing edge and background activations and yielding manipulation points.

\begin{table}[t!]
    \centering
    \captionsetup{font=small}
    \renewcommand{\arraystretch}{1.2}
    \setlength{\tabcolsep}{1.45mm}
    \caption{\small Success rates of affordance grounding manipulation in real-world scenarios .}
    \begin{tabular}{@{}lcccc@{}}
        \hline
        \textbf{Method} & \textbf{\textit{hold short-shirt}}  & \textbf{\textit{pull out tissue}}    & \textbf{\textit{hang mask}}        & \textbf{\textit{hang hat}} \\
        \hline \hline
        OS-AGDO~\cite{jia2025one} & 6 / 20  & 12 / 20   & 7 / 20        & 12 / 20 \\
        Ours & 10 / 20  & 16 / 20   & 12 / 20        & 17 / 20 \\
        \hline
    \end{tabular}
    \label{success_rate}
    \vskip-2ex
\end{table}

\textbf{Long-horizon Manipulation Demonstration.} 
The results in Fig.~\ref{process} illustrate the complete sequence from instruction decomposition to physical execution.
Based on visual input and TA-CoT, the system dynamically generates affordance semantic instructions and manipulation regions, where red markers denote grasping points and green markers denote placement points. In each inference cycle, the system outputs synchronized coordinates for both arms, achieving stable and continuous bimanual collaborative manipulation.

As shown in Table~\ref{success_rate}, the bimanual task ``Hold T-shirt'' is more challenging due to the coordination required between both arms. In contrast, the three single-arm tasks, ``Pull out tissue'', ``Hang mask'', and ``Hang hat'', achieve relatively high success rates, indicating stable grounding and execution across different object categories. Compared with OS-AGDO~\cite{jia2025one}, our method improves the average success rate from $46.25\%$ to $68.75\%$, corresponding to an average gain of $22.5$ percentage points across the four tasks, validating the robustness of the proposed framework in real-world manipulation.

\begin{figure}[!t]
  \centering
  \small
  \includegraphics[width=0.48\textwidth]{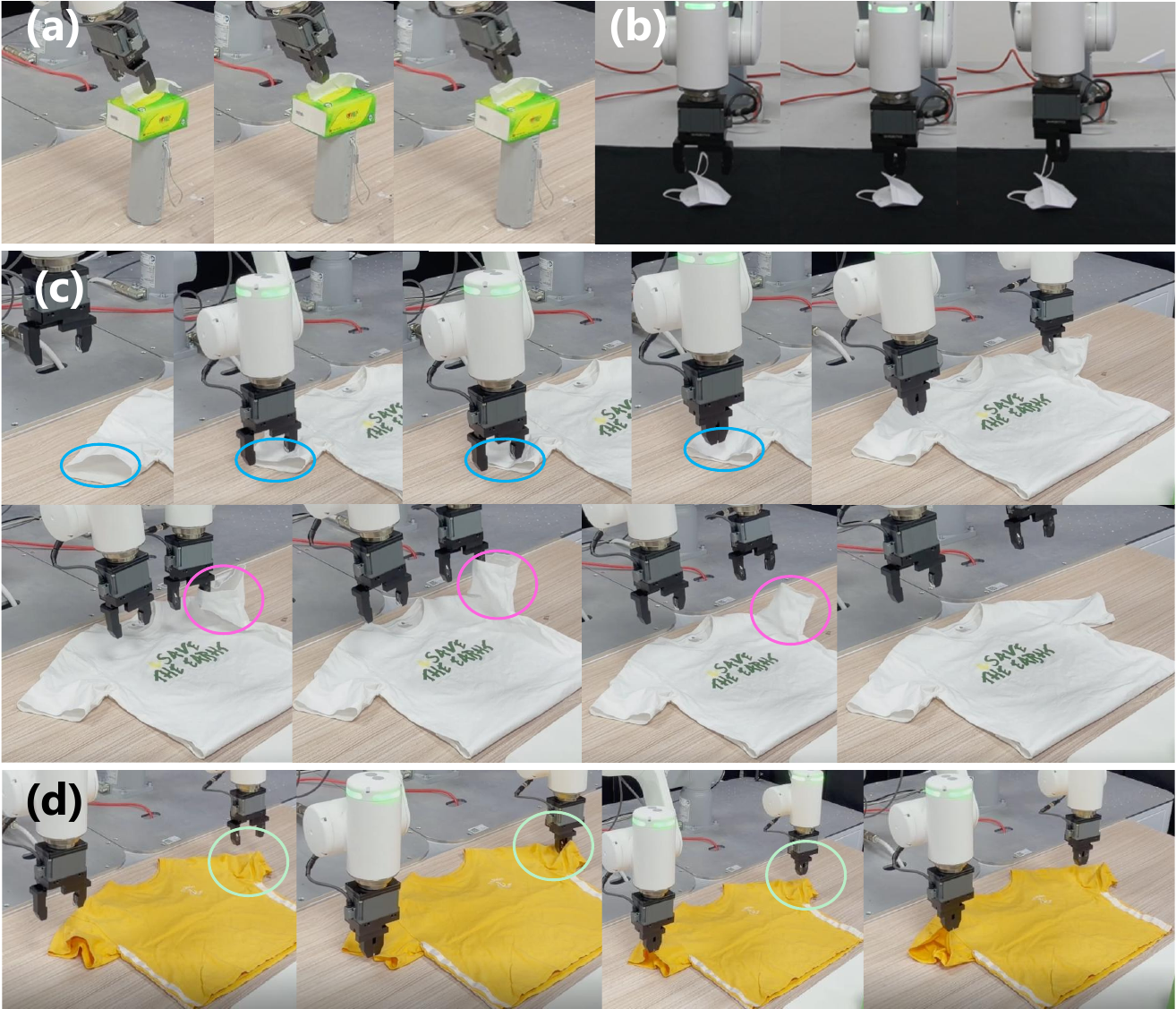}
  \captionsetup{font=small}
  \caption{\small Analysis of typical failure cases in real-world experiments.}
 \label{fail}
 \vskip-2ex
\end{figure}

\textbf{Failure Analysis and Future Directions.} 
Although the proposed method demonstrates improved manipulation capability for deformable objects, execution failures may still occur under challenging visual conditions, extreme object configurations, and physical constraints. To this end, we summarize the typical failure modes observed in real-world experiments, as illustrated in Fig.~\ref{fail}. First, for small, thin deformable objects, depth perception is highly sensitive to environmental noise, which can lead to inaccurate grasp planning and grasp failure, as shown in Fig.~\ref{fail}(a). Second, as shown in Fig.~\ref{fail}(b), the white mask suffers from strong light reflection, making visual recognition and affordance localization more difficult. Moreover, manipulating its thin straps requires finer-grained rope-like interaction, further increasing execution difficulty and reducing the success rate. Third, complex contact dynamics remain another major source of failure. Grasp slippage caused by local deformation at contact is indicated by the blue circle in Fig.~\ref{fail}(c), while the pink circle highlights the pull-back effect induced by frictional coupling when the object is grasped too deeply during retraction. Finally, some failures arise from hardware-level geometric constraints. As indicated by the green circle in Fig.~\ref{fail}(d), the current two-finger gripper struggles to establish stable grasps on extremely thin garments lying flat against the tabletop.

In summary, these failures reveal remaining limitations in visual perception under adverse lighting, depth estimation for thin objects, contact dynamics modeling, and end-effector adaptability, motivating future research on multimodal perception, contact-aware learning, and coordinated soft--hard design.

\section{Conclusion and Future Work}

In this work, we propose Texture-Robust Affordance Chain-of-thought for dEformable-object Region Grounding (TRACER), a long-horizon affordance grounding framework designed for complex-textured deformable objects. 
The system successfully establishes a cross-hierarchical mapping from high-level semantic reasoning to fine-grained physical interaction regions.
Specifically, the Tree-structured Affordance Chain-of-Thought (TA-CoT) decomposes abstract task intentions into hierarchical affordance-semantic instructions, providing highly interpretable semantic guidance for complex organization tasks.
To suppress prediction overflow beyond physical boundaries, the Spatially-Constrained Boundary Refinement (SCBR) loss effectively induces prediction contours to converge within the object’s interior. 
Simultaneously, to eliminate functional region fragmentation, the Interactive Convergent Refinement Flow (ICRF) module uses a learned acceleration field to refine dispersed responses, improving both spatial continuity and physical consistency.
Experimental results on the Fine-AGDDO15 dataset and a real-world dual-arm robotic platform demonstrate that TRACER achieves consistent improvements in localization precision and manipulation success rates under complex textural interference, effectively bridging the gap between semantic reasoning and robust physical execution.

To improve robustness, future work will focus on vision–tactile multimodal perception. Incorporating fine-grained force feedback will reduce depth uncertainty for extremely thin objects, mitigating current hardware geometric constraints. Furthermore, we plan to develop dynamic adaptive control strategies that leverage these tactile cues to ensure stable manipulation under complex, real-world interaction conditions.

\bibliographystyle{IEEEtran}   
\bibliography{ref} 

\begin{thebibliography}{10}
\providecommand{\url}[1]{#1}
\csname url@samestyle\endcsname
\providecommand{\newblock}{\relax}
\providecommand{\bibinfo}[2]{#2}
\providecommand{\BIBentrySTDinterwordspacing}{\spaceskip=0pt\relax}
\providecommand{\BIBentryALTinterwordstretchfactor}{4}
\providecommand{\BIBentryALTinterwordspacing}{\spaceskip=\fontdimen2\font plus
\BIBentryALTinterwordstretchfactor\fontdimen3\font minus \fontdimen4\font\relax}
\providecommand{\BIBforeignlanguage}[2]{{%
\expandafter\ifx\csname l@#1\endcsname\relax
\typeout{** WARNING: IEEEtran.bst: No hyphenation pattern has been}%
\typeout{** loaded for the language `#1'. Using the pattern for}%
\typeout{** the default language instead.}%
\else
\language=\csname l@#1\endcsname
\fi
#2}}
\providecommand{\BIBdecl}{\relax}
\BIBdecl

\bibitem{lu2024garmentlab}
H.~Lu \emph{et~al.}, ``{GarmentLab:} {A} unified simulation and benchmark for garment manipulation,'' in \emph{Proc. NeurIPS}, vol.~37, 2024, pp. 11\,866--11\,903.

\bibitem{borras2020grasping}
J.~Borras, G.~Alenya, and C.~Torras, ``A grasping-centered analysis for cloth manipulation,'' \emph{IEEE Transactions on Robotics}, vol.~36, no.~3, pp. 924--936, 2020.

\bibitem{wu2025garmentpile}
R.~Wu \emph{et~al.}, ``Garmentpile: Point-level visual affordance guided retrieval and adaptation for cluttered garments manipulation,'' in \emph{Proc. CVPR}, 2025, pp. 6950--6959.

\bibitem{zhou2025learning}
C.~Zhou \emph{et~al.}, ``Learning efficient robotic garment manipulation with standardization,'' in \emph{Proc. ICML}, 2025.

\bibitem{10966003}
------, ``{SSFold:} {Learning} to fold arbitrary crumpled cloth using graph dynamics from human demonstration,'' \emph{IEEE Transactions on Automation Science and Engineering}, vol.~22, pp. 14\,448--14\,460, 2025.

\bibitem{wu2024unigarmentmanip}
R.~Wu, H.~Lu, Y.~Wang, Y.~Wang, and H.~Dong, ``Unigarmentmanip: A unified framework for category-level garment manipulation via dense visual correspondence,'' in \emph{Proc. CVPR}, 2024, pp. 16\,340--16\,350.

\bibitem{wu2023learning}
R.~Wu, C.~Ning, and H.~Dong, ``Learning foresightful dense visual affordance for deformable object manipulation,'' in \emph{Proc. ICCV}, 2023, pp. 10\,913--10\,922.

\bibitem{zhang2024adaptigraph}
K.~Zhang, B.~Li, K.~Hauser, and Y.~Li, ``{AdaptiGraph:} {Material-adaptive} graph-based neural dynamics for robotic manipulation,'' in \emph{Proc. RSS}, 2024.

\bibitem{caporali2025gnn}
A.~Caporali, K.~Galassi, R.~Zanella, and G.~Palli, ``{GNN} topology representation learning for deformable multi-linear objects dual-arm robotic manipulation,'' \emph{IEEE Transactions on Automation Science and Engineering}, vol.~22, pp. 14\,738--14\,751, 2025.

\bibitem{zhuang2025flat}
L.~Zhuang, S.~Fan, Y.~Ru, F.~P. Audonnet, P.~Henderson, and G.~Aragon{-}Camarasa, ``{Flat'n'Fold:} {A} diverse multi-modal dataset for garment perception and manipulation,'' in \emph{Proc. ICRA}, 2025, pp. 7937--7944.

\bibitem{canberk2022cloth}
A.~Canberk \emph{et~al.}, ``Cloth funnels: Canonicalized-alignment for multi-purpose garment manipulation,'' in \emph{Proc. ICRA}, 2023, pp. 5872--5879.

\bibitem{yang2024task}
F.~Yang \emph{et~al.}, ``Task-oriented tool manipulation with robotic dexterous hands: A knowledge graph approach from fingers to functionality,'' \emph{IEEE Transactions on Cybernetics}, vol.~55, no.~1, pp. 395--408, 2025.

\bibitem{yang2025learning}
------, ``Learning granularity-aware affordances from human-object interaction for tool-based functional dexterous grasping,'' \emph{IEEE Transactions on Neural Networks and Learning Systems}, vol.~36, no.~11, pp. 19\,589--19\,603, 2025.

\bibitem{jia2025one}
W.~Jia \emph{et~al.}, ``One-shot affordance grounding of deformable objects in egocentric organizing scenes,'' in \emph{Proc. IROS}, 2025, pp. 21\,243--21\,250.

\bibitem{park2025trace}
S.~Park, J.~Kim, Y.~Cui, and M.~S. Brown, ``{TRACE:} {Textual} reasoning for affordance coordinate extraction,'' \emph{arXiv preprint arXiv:2511.01999}, 2025.

\bibitem{song2025robospatial}
C.~H. Song, V.~Blukis, J.~Tremblay, S.~Tyree, Y.~Su, and S.~Birchfield, ``{RoboSpatial:} {Teaching} spatial understanding to {2D} and {3D} vision-language models for robotics,'' in \emph{Proc. CVPR}, 2025, pp. 15\,768--15\,780.

\bibitem{clegg2020learning}
A.~Clegg, Z.~Erickson, P.~Grady, G.~Turk, C.~C. Kemp, and C.~K. Liu, ``Learning to collaborate from simulation for robot-assisted dressing,'' \emph{IEEE Robotics and Automation Letters}, vol.~5, no.~2, pp. 2746--2753, 2020.

\bibitem{sunil2025reactive}
N.~Sunil, M.~Tippur, A.~Saumell, E.~Adelson, and A.~Rodriguez, ``Reactive in-air clothing manipulation with confidence-aware dense correspondence and visuotactile affordance,'' in \emph{Proc. CoRL}, 2025.

\bibitem{tian2025diffusion}
T.~Tian, H.~Li, B.~Ai, X.~Yuan, Z.~Huang, and H.~Su, ``Diffusion dynamics models with generative state estimation for cloth manipulation,'' in \emph{Proc. CoRL}, 2025.

\bibitem{du2025physics}
Y.~Du, L.~Liu, and Z.~Zhang, ``A physics-informed neural network framework for real-time model predictive shape manipulation of deformable linear objects,'' \emph{IEEE Transactions on Automation Science and Engineering}, vol.~23, pp. 1666--1682, 2026.

\bibitem{kerr2025eye}
J.~Kerr \emph{et~al.}, ``Eye, robot: Learning to look to act with a {BC-RL} perception-action loop,'' in \emph{Proc. CoRL}, 2025.

\bibitem{pan2025omnimanip}
M.~Pan, J.~Zhang, T.~Wu, Y.~Zhao, W.~Gao, and H.~Dong, ``{OmniManip:} {Towards} general robotic manipulation via object-centric interaction primitives as spatial constraints,'' in \emph{Proc. CVPR}, 2025, pp. 17\,359--17\,369.

\bibitem{zhao2025cot}
Q.~Zhao \emph{et~al.}, ``{CoT-VLA:} {Visual} chain-of-thought reasoning for vision-language-action models,'' in \emph{Proc. CVPR}, 2025, pp. 1702--1713.

\bibitem{10602544}
M.~Zare, P.~M. Kebria, A.~Khosravi, and S.~Nahavandi, ``A survey of imitation learning: Algorithms, recent developments, and challenges,'' \emph{IEEE Transactions on Cybernetics}, vol.~54, no.~12, pp. 7173--7186, 2024.

\bibitem{huang2024rekep}
W.~Huang, C.~Wang, Y.~Li, R.~Zhang, and L.~Fei-Fei, ``{ReKep:} {Spatio-temporal} reasoning of relational keypoint constraints for robotic manipulation,'' in \emph{Proc. CoRL}, vol. 270, 2024, pp. 4573--4602.

\bibitem{chen2023learning}
W.~Chen, D.~Lee, D.~Chappell, and N.~Rojas, ``Learning to grasp clothing structural regions for garment manipulation tasks,'' in \emph{Proc. IROS}, 2023, pp. 4889--4895.

\bibitem{chen2025graphgarment}
W.~Chen, K.~Li, D.~Lee, X.~Chen, R.~Zong, and P.~Kormushev, ``{GraphGarment:} {Learning} garment dynamics for bimanual cloth manipulation tasks,'' \emph{arXiv preprint arXiv:2503.05817}, 2025.

\bibitem{zhang2022learning}
F.~Zhang and Y.~Demiris, ``Learning garment manipulation policies toward robot-assisted dressing,'' \emph{Science Robotics}, vol.~7, no.~65, p. eabm6010, 2022.

\bibitem{sun2024force}
Z.~Sun, Y.~Wang, D.~Held, and Z.~Erickson, ``Force-constrained visual policy: Safe robot-assisted dressing via multi-modal sensing,'' \emph{IEEE Robotics and Automation Letters}, vol.~9, no.~5, pp. 4178--4185, 2024.

\bibitem{weng2022fabricflownet}
T.~Weng, S.~M. Bajracharya, Y.~Wang, K.~Agrawal, and D.~Held, ``{FabricFlowNet:} {Bimanual} cloth manipulation with a flow-based policy,'' in \emph{Proc. CoRL}, vol. 164, 2021, pp. 192--202.

\bibitem{jiang2025phystwin}
H.~Jiang, H.-Y. Hsu, K.~Zhang, H.-N. Yu, S.~Wang, and Y.~Li, ``{PhysTwin:} {Physics-informed} reconstruction and simulation of deformable objects from videos,'' \emph{arXiv preprint arXiv:2503.17973}, 2025.

\bibitem{black2410pi0}
K.~Black \emph{et~al.}, ``{\(\pi\)}\({}_{\mbox{0}}\): {A} vision-language-action flow model for general robot control,'' \emph{arXiv preprint arXiv:2410.24164}, 2024.

\bibitem{intelligence2504pi05}
P.~Intelligence \emph{et~al.}, ``{\(\pi\)}\({}_{\mbox{0.5}}\): {a} vision-language-action model with open-world generalization,'' \emph{arXiv preprint arXiv:2504.16054}, 2025.

\bibitem{zhai2025igniting}
A.~Zhai \emph{et~al.}, ``Igniting {VLMs} toward the embodied space,'' \emph{arXiv preprint arXiv:2509.11766}, 2025.

\bibitem{hou2024key}
C.~Hou, Z.~Xue, B.~Zhou, J.~Ke, L.~Shao, and H.~Xu, ``Key-grid: Unsupervised {3D} keypoints detection using grid heatmap features,'' in \emph{Proc. NeurIPS}, vol.~37, 2024, pp. 49\,154--49\,179.

\bibitem{zhou2023clothesnet}
B.~Zhou \emph{et~al.}, ``{ClothesNet:} {An} information-rich {3D} garment model repository with simulated clothes environment,'' in \emph{Proc. ICCV}, 2023, pp. 20\,371--20\,381.

\bibitem{wang2025dexgarmentlab}
Y.~Wang \emph{et~al.}, ``{DexGarmentLab:} {Dexterous} garment manipulation environment with generalizable policy,'' in \emph{Proc. NeurIPS}, 2025.

\bibitem{zhao2025manipbench}
E.~Zhao \emph{et~al.}, ``{ManipBench:} {Benchmarking} vision-language models for low-level robot manipulation,'' in \emph{Proc. CoRL}, 2025.

\bibitem{ahn2022can}
M.~Ahn \emph{et~al.}, ``Do as {I} can, not as {I} say: {Grounding} language in robotic affordances,'' in \emph{Proc. CoRL}, vol. 205, 2022, pp. 287--318.

\bibitem{wang2025affordancer1}
H.~Wang \emph{et~al.}, ``{Affordance-R1:} {Reinforcement} learning for generalizable affordance reasoning in multimodal large language model,'' \emph{arXiv preprint arXiv:2508.06206}, 2025.

\bibitem{wu2025ragnet}
D.~Wu \emph{et~al.}, ``{RAGNet:} {Large-scale} reasoning-based affordance segmentation benchmark towards general grasping,'' in \emph{Proc. ICCV}, 2025, pp. 11\,980--11\,990.

\bibitem{jose2024dinov2meetstextunified}
C.~Jose \emph{et~al.}, ``{DINOv2} meets text: {A} unified framework for image- and pixel-level vision-language alignment,'' in \emph{Proc. CVPR}, 2024, pp. 24\,905--24\,916.

\bibitem{gao2025vita}
D.~Gao \emph{et~al.}, ``{VITA:} {Vision-to-action} flow matching policy,'' \emph{arXiv preprint arXiv:2507.13231}, 2025.

\bibitem{liu2022flow}
X.~Liu, C.~Gong, and Q.~Liu, ``Flow straight and fast: Learning to generate and transfer data with rectified flow,'' in \emph{Proc. ICLR}, 2023.

\bibitem{noobject}
F.~Li, X.~Wang, X.~Wang, Z.~Zhang, and Y.~Xu, ``No object is an island: Enhancing {3D} semantic segmentation generalization with diffusion models,'' in \emph{Proc. NeurIPS}, 2025.

\bibitem{zhang2025towards}
Y.~Zhang, Y.~Yan, A.~G. Schwing, and Z.~Zhao, ``Towards hierarchical rectified flow,'' in \emph{Proc. ICLR}, 2025.

\bibitem{wei2025afforddexgrasp}
Y.-L. Wei \emph{et~al.}, ``{AffordDexGrasp:} {Open-set} language-guided dexterous grasp with generalizable-instructive affordance,'' in \emph{Proc. ICCV}, 2025, pp. 11\,818--11\,828.

\bibitem{zhang2025h2oflow}
H.~Zhang and L.~Carlone, ``{H2OFlow:} {Grounding} human-object affordances with {3D} generative models and dense diffused flows,'' \emph{arXiv preprint arXiv:2510.21769}, 2025.

\bibitem{Kirillov_2023_ICCV}
A.~Kirillov \emph{et~al.}, ``Segment anything,'' in \emph{Proc. ICCV}, 2023, pp. 3992--4003.

\bibitem{li2024one}
G.~Li, D.~Sun, L.~Sevilla-Lara, and V.~Jampani, ``One-shot open affordance learning with foundation models,'' in \emph{Proc. CVPR}, 2024, pp. 3086--3096.

\bibitem{chen2024internvl}
Z.~Chen \emph{et~al.}, ``How far are we to {GPT-4V?} {Closing} the gap to commercial multimodal models with open-source suites,'' \emph{Science China Information Sciences}, vol.~67, no.~12, p. 220101, 2024.

\bibitem{hurst2024gpt}
A.~Hurst \emph{et~al.}, ``{GPT-4o} system card,'' \emph{arXiv preprint arXiv:2410.21276}, 2024.

\bibitem{comanici2025gemini}
G.~Comanici \emph{et~al.}, ``Gemini 2.5: Pushing the frontier with advanced reasoning, multimodality, long context, and next generation agentic capabilities,'' \emph{arXiv preprint arXiv:2507.06261}, 2025.

\end{thebibliography}
\end{document}